\documentclass[Afour,sageh,times]{sagej}

\usepackage{moreverb,url}

\usepackage[colorlinks,bookmarksopen,bookmarksnumbered,citecolor=red,urlcolor=red]{hyperref}

\usepackage[title,toc,page]{appendix}

\usepackage{enumitem}

\usepackage[printonlyused]{acronym}

\usepackage{siunitx}
\usepackage[all]{nowidow}

\usepackage[table]{xcolor}

\usepackage{tikz}
\usetikzlibrary{arrows}
\usetikzlibrary{arrows.meta}
\usepackage{makecell}
\usetikzlibrary{positioning}
\usetikzlibrary{decorations.pathreplacing,calligraphy}
\usetikzlibrary{calc}
\usepackage[normalem]{ulem}
\usepackage{lipsum}

\usepackage{hhline}

\usepackage{xspace} 

\usepackage{pifont}
\newcommand{\cmark}{\ding{51}}%
\newcommand{\xmark}{\ding{55}}%

\newcommand{\rotheadsixty}[1]{\rotatebox{57}{#1}}

\usepackage{epstopdf}

\usepackage{import}

\usepackage{booktabs}
\usepackage{array}
\usepackage{longtable}
\newcolumntype{L}{>{\raggedright\arraybackslash}p{0.13\textwidth}}
\newcolumntype{C}{>{\centering\arraybackslash}p{0.10\textwidth}}
\newcolumntype{M}{>{\centering\arraybackslash}p{0.065\textwidth}}

\usepackage{tabularx}
\usepackage{multirow, multicol}

\newlength{\Oldarrayrulewidth}

\newcolumntype{?}[1]{!{\vrule width #1}}

\usepackage{amssymb,amsfonts,amsmath,amscd}

\usepackage{bm}

\usepackage{balance}
\usepackage[T1]{fontenc}

\newcommand{\rev}[1]{{\color{black}{#1}}}
\usepackage{cancel}

\newcommand{\bbm}{\begin{bmatrix}}
\newcommand{\ebm}{\end{bmatrix}}

\usepackage{float}

\newcommand{\ignore}[1]{}

\newcommand{\bma}[1]{\left[\begin{array}{#1}}
\newcommand{\ema}{\end{array}\right]}

\DeclareMathAlphabet{\mbf}{OT1}{ptm}{b}{n}
\newcommand{\mbs}[1]{{\boldsymbol{#1}}}

\def\fdotb{{\raisebox{-0.6ex}{ \kern0.2ex\raisebox{0.8ex}{\tiny $\hspace*{-1ex}\circ$}}}}
\def\fddotb{{\raisebox{-0.6ex}{ \kern0.2ex\raisebox{0.8ex}{\tiny $\hspace*{-1ex}\circ\circ$}}}}

\newcommand{\trans}{{\ensuremath{\mathsf{T}}}} 
\newcommand{\utimes}{ {\raisebox{-0.6ex}{ \kern-1.0ex\raisebox{0.6ex}{ \small $\mathsf{v}$}}} } %
\newcommand{\beq}{\begin{equation}}
\newcommand{\eeq}{\end{equation}}
\newcommand{\bdis}{\begin{displaymath}}
\newcommand{\edis}{\end{displaymath}}
\newcommand{\beqarray}{\begin{eqnarray}}
\newcommand{\eeqarray}{\end{eqnarray}}
\newcommand{\beqarraynn}{\begin{eqnarray*}}
\newcommand{\eeqarraynn}{\end{eqnarray*}}

\newcommand{\cross}{\times}

\DeclareMathAlphabet{\mbf}{OT1}{ptm}{b}{n}

\usepackage{hyperref}

\newcommand\BibTeX{{\rmfamily B\kern-.05em \textsc{i\kern-.025em b}\kern-.08em
T\kern-.1667em\lower.7ex\hbox{E}\kern-.125emX}}

\begin{document}

\runninghead{Lisus et al.}

\title{Boreas Road Trip: A Multi-Sensor Autonomous Driving Dataset on Challenging Roads}

\author{Daniil Lisus\affilnum{1}, Katya M. Papais \affilnum{1}, Cedric Le Gentil\affilnum{1}, Elliot Preston-Krebs\affilnum{1}, Andrew Lambert\affilnum{2}, Keith Y.K. Leung\affilnum{2}, and Timothy D. Barfoot\affilnum{1}}

\affiliation{\affilnum{1}University of Toronto Institute of Aerospace Studies, Toronto, Canada\\
\affilnum{2}Trimble, Canada}

\corrauth{Daniil Lisus}

\email{daniil.lisus@robotics.utias.utoronto.ca}

\begin{abstract}
The Boreas Road Trip (Boreas-RT) dataset extends the multi-season Boreas dataset to new and diverse locations that pose challenges for modern autonomous driving algorithms.
Boreas-RT comprises 60 sequences collected over 9 real-world routes, totalling 643 km of driving.
Each route is traversed multiple times, enabling evaluation in identical environments under varying traffic and, in some cases, weather conditions.
The data collection platform includes a 5MP FLIR Blackfly S camera, a 360\si{\degree} Navtech RAS6 Doppler-enabled spinning radar, a 128-channel 360\si{\degree} Velodyne Alpha Prime lidar, an Aeva Aeries~II FMCW Doppler-enabled lidar, a Silicon Sensing DMU41 inertial measurement unit, and a Dynapar wheel encoder.
Centimetre-level ground truth is provided via post-processed Applanix POS LV GNSS-INS data.
The dataset includes precise extrinsic and intrinsic calibrations, a publicly available development kit, and a live leaderboard for odometry and metric localization.
Benchmark results show that many state-of-the-art odometry and localization algorithms overfit to simple driving environments and degrade significantly on the more challenging Boreas-RT routes. Boreas-RT provides a unified dataset for evaluating multi-modal algorithms across diverse road conditions.
The dataset, leaderboard, and development kit are available at \href{https://www.boreas.utias.utoronto.ca/#/boreasRT}{boreas.utias.utoronto.ca}.
\end{abstract}

\keywords{Autonomous vehicles, dataset, camera, radar, lidar, FMCW lidar, IMU, GPS}

\maketitle

\section{Introduction}

Over the past decade, the autonomous vehicle (AV) research community has seen large performance gains across many tasks.
These gains are partly driven by the proliferation of real-world datasets, which enable direct and fair algorithmic comparison.
Over time, such comparisons signal promising research directions and help identify the most impactful methodological advances.
However, the high cost of data collection imposes a trade-off: datasets typically prioritize either diversity (\textit{breadth}) or repeated coverage (\textit{depth}).

\begin{figure}[t]
    \centering
    \leavevmode\scalebox{0.95}{
        \includegraphics[width=0.50\textwidth, trim={0.2cm 0cm 0 0cm},clip]{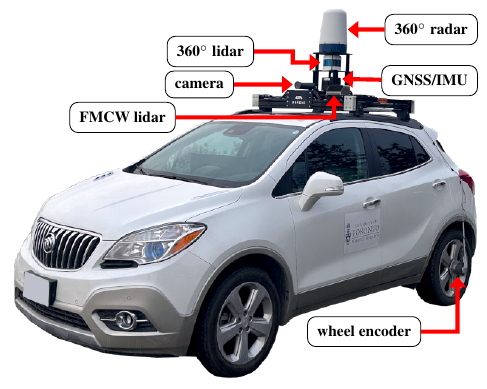}
    }
    \caption{Our data collection platform, \textit{Boreas}, is equipped with a 5MP FLIR Blackfly S camera, a 360\si{\degree} Navtech RAS6 Doppler-enabled spinning radar, a 360\si{\degree} Velodyne Alpha Prime lidar, an Aeva Aeries~II FMCW Doppler-enabled lidar, a Silicon Sensing DMU41 IMU, and a Dynapar wheel encoder.}
    \label{fig:boreas}
\end{figure}

\renewcommand{\arraystretch}{1.2}
\begin{table*}[ht]
\centering
\caption{Related datasets intended for odometry, mapping, or localization algorithm development. Lead: Public leaderboard. GT: Ground truth pose source. INS: GNSS and IMU system. RTK: Real-Time Kinematic INS approach that uses a GNSS base station to correct an on-board GNSS. RTX: Real-Time eXtended INS approach that leverages a network of GNSS base stations to correct an on-board GNSS (see \cite{applanix2022}). SLAM: Simultaneous localization and mapping. VO: Visual odometry. F-lidar: FMCW lidar.}
\label{tab:rw}
\begin{tabularx}{\linewidth}{cccccccccccc}
\toprule
name & lead & length & GT & \rotheadsixty{camera} & \rotheadsixty{lidar} & 
\rotheadsixty{f-lidar} & \rotheadsixty{radar} & \rotheadsixty{urban} & 
\rotheadsixty{suburban} & \rotheadsixty{rural} & \rotheadsixty{highway}\\
\midrule

KITTI (Odometry) &  & $39\; \si{\km}$ &  &  &  &  &  &  &  &  & \\[-2.2pt]
\cite{geiger2012kitti} & \multirow{-2}{*}{\cmark} & $22$ seqs & \multirow{-2}{*}{INS+RTK} &
\multirow{-2}{*}{\cmark} & \multirow{-2}{*}{\cmark} & \multirow{-2}{*}{\xmark} &
\multirow{-2}{*}{\xmark} & \multirow{-2}{*}{\cmark} & \multirow{-2}{*}{\cmark} &
\multirow{-2}{*}{\xmark} & \multirow{-2}{*}{\xmark} \\

\rowcolor{gray!10}
Oxford RobotCar &  & $1000 \; \si{\km}$ &  &  &  &  &  &  &  &  & \\[-2.2pt]
\rowcolor{gray!10}
\cite{oxford_robotcar} & \multirow{-2}{*}{\xmark} & $100$ seqs & \multirow{-2}{*}{INS+RTK} &
\multirow{-2}{*}{\cmark} & \multirow{-2}{*}{\cmark} & \multirow{-2}{*}{\xmark} &
\multirow{-2}{*}{\xmark} & \multirow{-2}{*}{\cmark} & \multirow{-2}{*}{\xmark} &
\multirow{-2}{*}{\xmark} & \multirow{-2}{*}{\xmark} \\

Complex Urban &  & $451 \; \si{\km}$ &  &  &  &  &  &  &  &  & \\[-2.2pt]
\cite{complex_urban} & \multirow{-2}{*}{\xmark} & $40$ seqs & \multirow{-2}{*}{SLAM} &
\multirow{-2}{*}{\cmark} & \multirow{-2}{*}{\cmark} & \multirow{-2}{*}{\xmark} &
\multirow{-2}{*}{\xmark} & \multirow{-2}{*}{\cmark} & \multirow{-2}{*}{\xmark} &
\multirow{-2}{*}{\xmark} & \multirow{-2}{*}{\xmark} \\

\rowcolor{gray!10}
UrbanLoco &  & $40 \; \si{\km}$ &  &  &  &  &  &  &  &  & \\[-2.2pt]
\rowcolor{gray!10}
\cite{urbanloco} & \multirow{-2}{*}{\xmark} & $13$ seqs & \multirow{-2}{*}{INS+RTK} &
\multirow{-2}{*}{\cmark} & \multirow{-2}{*}{\cmark} & \multirow{-2}{*}{\xmark} &
\multirow{-2}{*}{\xmark} & \multirow{-2}{*}{\cmark} & \multirow{-2}{*}{\xmark} &
\multirow{-2}{*}{\xmark} & \multirow{-2}{*}{\xmark} \\

MulRan &  & $124 \; \si{\km}$ &  &  &  &  &  &  &  &  & \\[-2.2pt]
\cite{kim2020mulran} & \multirow{-2}{*}{\xmark} & $12$ seqs & \multirow{-2}{*}{SLAM} &
\multirow{-2}{*}{\xmark} & \multirow{-2}{*}{\cmark} & \multirow{-2}{*}{\xmark} &
\multirow{-2}{*}{\cmark} & \multirow{-2}{*}{\cmark} & \multirow{-2}{*}{\xmark} &
\multirow{-2}{*}{\xmark} & \multirow{-2}{*}{\xmark} \\

\rowcolor{gray!10}
Oxford Radar RobotCar &  & $280 \; \si{\km}$ &  &  &  &  &  &  &  &  & \\[-2.2pt]
\rowcolor{gray!10}
\cite{oxford_radar_robotcar} & \multirow{-2}{*}{\xmark} & $32$ seqs & \multirow{-2}{*}{INS+VO} &
\multirow{-2}{*}{\cmark} & \multirow{-2}{*}{\cmark} & \multirow{-2}{*}{\xmark} &
\multirow{-2}{*}{\cmark} & \multirow{-2}{*}{\cmark} & \multirow{-2}{*}{\xmark} &
\multirow{-2}{*}{\xmark} & \multirow{-2}{*}{\xmark} \\

Boreas &  & $350 \; \si{\km}$ &  &  &  &  &  &  &  &  & \\[-2.2pt]
\cite{burnett2023boreas} & \multirow{-2}{*}{\cmark} & $44$ seqs & \multirow{-2}{*}{INS+RTX} &
\multirow{-2}{*}{\cmark} & \multirow{-2}{*}{\cmark} & \multirow{-2}{*}{\xmark} &
\multirow{-2}{*}{\cmark} & \multirow{-2}{*}{\cmark} & \multirow{-2}{*}{\cmark} &
\multirow{-2}{*}{\xmark} & \multirow{-2}{*}{\xmark} \\

\rowcolor{gray!10}
Oxford Offroad Radar &  & $154 \; \si{\km}$ &  &  &  &  &  &  &  &  & \\[-2.2pt]
\rowcolor{gray!10}
\cite{oord_dataset} & \multirow{-2}{*}{\xmark} & $11$ seqs & \multirow{-2}{*}{INS} &
\multirow{-2}{*}{\xmark} & \multirow{-2}{*}{\xmark} & \multirow{-2}{*}{\xmark} &
\multirow{-2}{*}{\cmark} & \multirow{-2}{*}{\xmark} & \multirow{-2}{*}{\xmark} &
\multirow{-2}{*}{\cmark} & \multirow{-2}{*}{\xmark} \\

HeLiPR &  & $164 \; \si{\km}$ &  &  &  &  &  &  &  &  & \\[-2.2pt]
\cite{helipr} & \multirow{-2}{*}{\xmark} & $10$ seqs & \multirow{-2}{*}{INS} &
\multirow{-2}{*}{\xmark} & \multirow{-2}{*}{\cmark} & \multirow{-2}{*}{\cmark} &
\multirow{-2}{*}{\xmark} & \multirow{-2}{*}{\cmark} & \multirow{-2}{*}{\xmark} &
\multirow{-2}{*}{\xmark} & \multirow{-2}{*}{\cmark} \\

\rowcolor{gray!10}
HeRCULES &  & $59 \; \si{\km}$ &  &  &  &  &  &  &  &  & \\[-2.2pt]
\rowcolor{gray!10}
\cite{hercules} & \multirow{-2}{*}{\xmark} & $21$ seqs & \multirow{-2}{*}{INS+RTK} &
\multirow{-2}{*}{\cmark} & \multirow{-2}{*}{\xmark} & \multirow{-2}{*}{\cmark} &
\multirow{-2}{*}{\cmark} & \multirow{-2}{*}{\cmark} & \multirow{-2}{*}{\xmark} &
\multirow{-2}{*}{\xmark} & \multirow{-2}{*}{\cmark} \\

 &  & $643 \; \si{\km}$ &  &  &  &  &  &  &  &  & \\[-2.2pt]
\multirow{-2}{*}{\textbf{Boreas-RT}} & \multirow{-2}{*}{\cmark} & $60$ seqs & \multirow{-2}{*}{INS+RTK} &
\multirow{-2}{*}{\cmark} & \multirow{-2}{*}{\cmark} & \multirow{-2}{*}{\cmark} &
\multirow{-2}{*}{\cmark} & \multirow{-2}{*}{\cmark} & \multirow{-2}{*}{\cmark} &
\multirow{-2}{*}{\cmark} & \multirow{-2}{*}{\cmark} \\

\bottomrule
\end{tabularx}
\end{table*}

Although recent perception datasets (e.g., \cite{kitti360, waymo_motion1, ApolloScape}) have begun achieving both diversity and scale, datasets designed for less learning-dependent AV \emph{state-estimation} tasks, such as odometry, mapping, and localization, remain largely split between diversity and repeatability.
For these tasks, the \textit{breadth} approaches enable testing across a wide range of road types, but ever-changing conditions (traffic, time of day, weather, etc.) make it difficult to disentangle the effects of environment type from transient conditions.
Conversely, the \textit{depth} approaches enable detailed study of how varying conditions affect performance, but increase the risk of overfitting to a particular road segment.
Compounding this issue, many `classic' state-estimation datasets (e.g., \cite{geiger2012kitti, oxford_robotcar}) lack several sensors found in modern AV stacks.
As a result, researchers often mix and match multiple datasets to support their claims, complicating development and reducing the ability to conduct rigorous one-to-one comparisons.

\rev{The Boreas Road Trip (Boreas-RT) dataset is intended to provide both \textit{breadth} and \textit{depth} for odometry, mapping, and localization tasks.
Boreas-RT extends the multi-season Boreas dataset (\citet{burnett2023boreas}) by including
\begin{itemize}[itemsep=0pt, parsep=0pt]
    \item eight new routes (nine total), 
    \item data from a stand-alone IMU that is not used in ground truth generation,
    \item data from a novel FMCW lidar, and
    \item data from the same 360$^\circ$ radar, but now with a new firmware that enables Doppler velocity extraction (\cite{lisus2025doppler}).
\end{itemize}
}
Each route is repeated on average six times for a total of $643\;\si{km}$ of driving data.
Traversals span different times and, in some cases, weather conditions to capture natural variations in road conditions.
The routes were chosen to challenge the targeted AV tasks and provide diversity in geometry (suburbs vs. long tunnels), speed (industrial areas vs. major freeways), and urbanization (forest roads vs. urban centres).
Boreas-RT includes a rich suite of sensors: a camera, a 360\si{\degree} Doppler-enabled spinning radar, a 360\si{\degree} lidar, an FMCW Doppler-enabled lidar\footnotemark \,with a built-in inertial measurement unit (IMU), a stand-alone IMU, and a wheel encoder.
Figure~\ref{fig:boreas} shows the data collection platform with all sensors.
The dataset provides centimetre-level post-processed GNSS–INS ground truth poses and millimetre-per-second-level ground truth velocities.
This ground truth is generated without using any of the raw sensor data (except the wheel encoder), ensuring that algorithm evaluations remain objective and free of correlations between the sensing modality under test and the ground truth.
This enables fair comparison across sensing modalities, road types, and conditions over multiple trials using a consistent data format.
We provide benchmarks showing where current state-of-the-art (SOTA) algorithms fail using the new data.

\footnotetext{The FMCW lidar is only present in half of the routes.}
\begin{table}[ht]
\centering
\caption{Boreas-RT sensor specifications. HFOV: Horizontal field of view. VFOV: Vertical field of view. BI: Bias instability. RW: Random walk. $^\dagger$Accuracy changes over time as a function of visible satellites and multi-path conditions with nominal performance reported here.}
\label{tab:sensors}
\begin{tabularx}{\linewidth}{p{1.85cm} X}
\toprule
{Sensor} & {Specifications} \\
\midrule

\rowcolor{gray!10}
Applanix & $\bullet$ 2--4 \si{\cm} absolute RTX accuracy (RMS)$^\dagger$\\
\rowcolor{gray!10}
\vspace{-13.5pt} POS LV 220 & $\bullet$ 5--10 \si{\mm/\second} velocity accuracy (RMS)$^\dagger$\\
\rowcolor{gray!10}
\vspace{-17.5pt} (GNSS-INS) & $\bullet$ 200 \si{\hertz}\\

FLIR & $\bullet$ 81\si{\degree} HFOV $\times$ 71\si{\degree} VFOV\\
\vspace{-13.5pt}Blackfly S & $\bullet$ 2448$\times$2048 (5MP)\\
\vspace{-17.5pt}Camera & $\bullet$ 10 \si{\hertz}\\
\vspace{-23pt}\scriptsize(BFS-U3-51S5C) & \\[-13pt]

\rowcolor{gray!10}
Navtech & $\bullet$ 360\si{\degree} HFOV\\
\rowcolor{gray!10}
\vspace{-13.5pt} RAS6 Radar & $\bullet$ 0.0438 \si{\m} range resolution\\
\rowcolor{gray!10}
& $\bullet$ 0.9\si{\degree} angular resolution \\
\rowcolor{gray!10}
& $\bullet$ 300 \si{\m} range \\
\rowcolor{gray!10}
& $\bullet$ 4 \si{\hertz} \\
\rowcolor{gray!10}
& $\bullet$ Doppler-enabled \citep{lisus2025doppler} \\

Velodyne & $\bullet$ 360\si{\degree} HFOV $\times$ 40\si{\degree} VFOV\\
\vspace{-14.5pt}Alpha-Prime & $\bullet$ 128 beams\\
\vspace{-17.5pt}Lidar & $\bullet$ 245 \si{\m} range\\
& $\bullet$ $\approx$ 2.2M points/s \\
& $\bullet$ 10 \si{\hertz} \\

\rowcolor{gray!10}
Aeva & $\bullet$ 120\si{\degree} HFOV $\times$ 30\si{\degree} VFOV\\
\rowcolor{gray!10}
\vspace{-14.5pt}Aeries II & $\bullet$ 500 \si{\m} range\\
\rowcolor{gray!10}
\vspace{-17.5pt}FMCW & $\bullet$ $\approx$ 1.0M points/s\\
\rowcolor{gray!10}
\vspace{-21.5pt}Lidar & $\bullet$ 10 \si{\hertz} \\
\rowcolor{gray!10}
& $\bullet$ 100 \si{\hertz} InvenSense IMU ({\scriptsize IAM-20680HP})\\
\rowcolor{gray!10}
& $\bullet$ Provides Doppler velocities \\

Silicon & $\bullet$ Angular BI: 0.1 \si{\degree/\hour}, RW: 0.02 \si{\degree/\sqrt{\hour}}\\
\vspace{-14.5pt}Sensing & $\bullet$ Linear BI: 15 \si{\micro g}, RW: 0.05 \si{\m/\second/\sqrt{\hour}}\\
\vspace{-17.5pt}DMU41 & $\bullet$ Range: $\pm 490 \si{\degree/\second}$ (ang.), $\pm 10 \si{g}$ (lin.)\\
\vspace{-19.5pt}IMU & $\bullet$ 200 \si{\hertz} \\

\rowcolor{gray!10}
Dynapar & $\bullet$ Incremental optical encoder \\
\rowcolor{gray!10}
\vspace{-13.5pt} Encoder & $\bullet$ 1024 pulses per revolution (PPR) \\
\rowcolor{gray!10}
\vspace{-17.5pt} \scriptsize(HS35R) & $\bullet$ Mounted to rear left wheel\\
\bottomrule
\end{tabularx}
\end{table}

\section{Related work}
A large number of datasets have been released over the past two decades targeting various parts of the AV stack.
A recently-compiled extensive list and comparison can be found in \cite{survey_av_datasets}.
In this paper, we restrict our comparison to datasets aligned with our target applications of AV odometry, mapping, and localization.
Such datasets feature distinct trajectories, provide ground truth poses over time, and collect data on roads intended for cars, even when those roads are not easily accessible.
Table~\ref{tab:rw} provides a summary of relevant datasets and how they compare to ours.

Our dataset provides the second-largest kilometre count among comparable datasets, second only to the Oxford RobotCar dataset \citep{oxford_robotcar}, which collects all 100 of its sequences on a single route, in contrast to our nine routes.
Boreas-RT is also one of the few datasets, alongside the Boreas \citep{burnett2023boreas}, Oxford Radar RobotCar \citep{oxford_radar_robotcar}, and HeRCULES \cite{hercules} datasets, to provide camera, lidar, and radar exteroceptive sensor data.
This enables evaluation of multi-sensor algorithms as well as comparison of individual sensing modalities.
Crucially, Boreas-RT is the first dataset to release FMCW lidar data alongside a full modern AV stack of sensors (camera, $360\;\si{\degree}$ lidar and radar).
This relatively new type of lidar provides Doppler velocity measurements for each returned point, similar to automotive radar \citep{pierrottet2008linear, aevascenes}.
Such relative velocity information has been shown to help mitigate geometric degeneracy \citep{Hexsel_Vhavle_Chen_2022, picking_up_speed} and improve odometry performance \citep{need_for_speed}, among other potential applications.
We also release centimetre-level ground truth poses from INS-RTK post-processing.
These poses are uncorrelated with the raw sensor data (aside from wheel-encoder measurements), enabling precise and fair algorithm evaluation.

A key motivation for creating Boreas-RT is to help prevent algorithmic overfitting to specific road types.
In support of this, Boreas-RT is the only dataset to feature roads spanning a wide range of urbanization levels.
While most datasets focus on urban and occasionally suburban routes, Boreas-RT includes sequences collected on diverse highways and rural roads.
Finally, our dataset is among the few to offer a public leaderboard, facilitating comparisons between different approaches.

\begin{figure}
    \centering
    \includegraphics[width=0.85\linewidth]{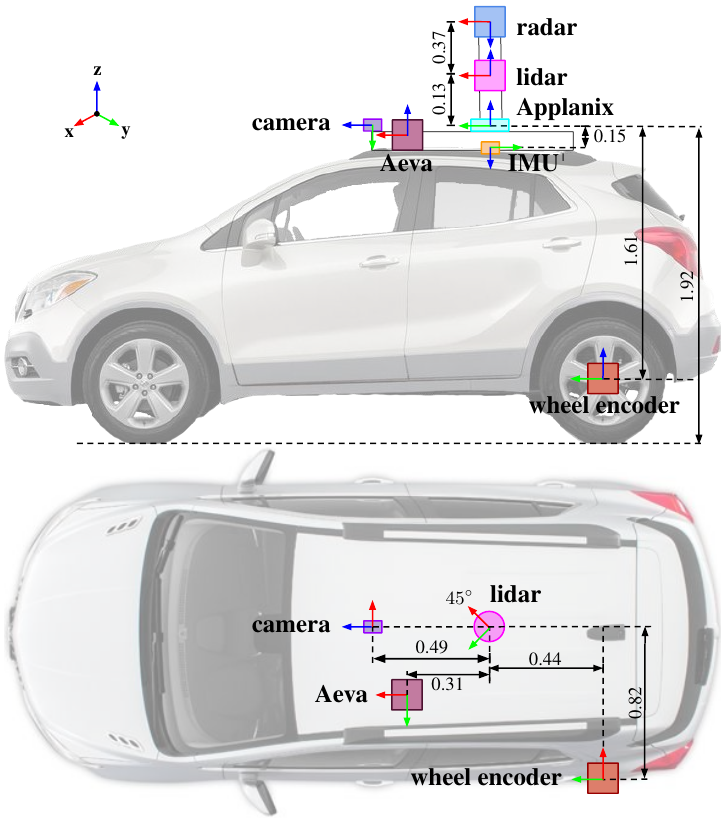}
    \caption{Boreas-RT sensor placements. All distances are given in metres, with exact values provided in the calibration folder. `Lidar' refers to the $360\;\si{\degree}$ Velodyne Alpha-Prime lidar, whereas `Aeva' refers to the Aeva Aeries II FMCW lidar.}
    \label{fig:sensors}
\end{figure}

\renewcommand{\arraystretch}{1.2}
\begin{table*}[t]
\centering
\caption{Overview of the different routes collected for Boreas-RT. Loop sequences start and end at the same location, while one-way sequences alternate start and end points. GT: The approximate ground truth per-axis repeatability accuracy between sequences as discussed in the Ground truth section. $^{\dagger}$\rev{All orientation repeatability errors are between $0.03^{\circ}$ and $0.06^{\circ}$}. $^\star$The Aeva lidar was mounted for only half of the \texttt{suburbs} sequences.}
\label{tab:sequences}
\begin{tabularx}{\linewidth}{ccccccccX}
\toprule
subset & route &  num seq. & seq. type & \multicolumn{1}{c}{\makecell{avg.\ length\\{}[\si{\km}]}} & \multicolumn{1}{c}{\makecell{avg. speed \\{} [\si{\km/\hour}]}} & \multicolumn{1}{c}{\makecell{GT$^{\dagger}$ \\ {}[\si{\meter}]}} & Aeva & challenges\\
\midrule

\rowcolor{gray!10}
 & \texttt{suburbs} & 10 & loop & 7.9 & 26.8 & $\leq 0.03$ & ~~\cmark$^\star$ & snow banks, snow \\
\rowcolor{gray!10}
& \texttt{industrial} & 5 & loop & 5.4 & 22.1 & $\leq 0.03$ & \cmark & snow banks, snowstorm \\
\rowcolor{gray!10}
\multirow{-3}{*}{\rotatebox[origin=c]{90}{structured}} & \texttt{urban} & 7 & loop & 8.6 & 13.2 & $\leq 0.80$ &  & many pedestrians \\

 & \texttt{forest} & 4 & loop & 16.4 & 52.0 & $\leq 0.40$ & & dense foliage \\
\multirow{-2}{*}{\rotatebox[origin=c]{90}{rural}} & \texttt{farm} & 10 & loop & 10.8 & 38.2 & $\leq 0.03$ & & dust clouds, featureless \\

\rowcolor{gray!10}
 & \texttt{tunnel} & 10 & one-way & 1.9 & 33.5 & $\leq 0.15$ &  \cmark & featureless \\
\rowcolor{gray!10}
& \texttt{skyway} & 5 & loop & 11.1 & 66.7 & $\leq 0.03$ &  \cmark & featureless \\
\rowcolor{gray!10}
& \texttt{regional} & 6 & one-way & 9.3 & 39.7 & $\leq 0.03$ & \cmark & varying levels of traffic \\
\rowcolor{gray!10}
\multirow{-4}{*}{\rotatebox[origin=c]{90}{highway}} & \texttt{freeway} & 3 & one-way & 57.6 & 61.8 & $\leq 0.03$ & & long continuous driving\\

\bottomrule
\multicolumn{2}{c}{Totals/Averages} & 60 &  & 10.7 & 35.9\\
\bottomrule
\end{tabularx}
\end{table*}

\section{Sensors}
All sensor specifications are listed in Table~\ref{tab:sensors}, and their relative placements on the vehicle are shown in Figure~\ref{fig:sensors}.

The Aeva Doppler-enabled FMCW lidar, referred to as `Aeva' or `FMCW lidar’ throughout, was mounted on the vehicle for only 31 of the 60 sequences.
In contrast, the Velodyne Alpha-Prime lidar, referred to as `Velodyne' or simply `lidar', was present for all 60 sequences.

Three types of IMU measurements are included: the primary stand-alone DMU41 IMU, the internal Aeva IMU (only present for sequences that contain Aeva data), and the internal Applanix POS LV IMU.
Note that the Applanix IMU is only provided for backward-compatibility with the original Boreas dataset, as it was the only IMU present on the platform at that time.
This IMU is used in the post-processing of ground truth data; therefore, its use should be avoided in algorithms to prevent correlation between algorithm estimates and the ground truth.
The presence of a truly stand-alone IMU is rare in state-estimation datasets and is valuable for testing algorithms that wish to leverage inertial measurements in a truly `probabilistically fair' way.

The wheel encoder is included as a stand-alone sensor to provide an alternative source of proprioceptive information.
It should be noted that the wheel encoder is used in the post-processing of ground truth data and so the use of it in an algorithm will make the algorithm estimates (slightly) correlated with the ground truth.

\begin{figure}[t]
    \centering
    \includegraphics[width=\linewidth]{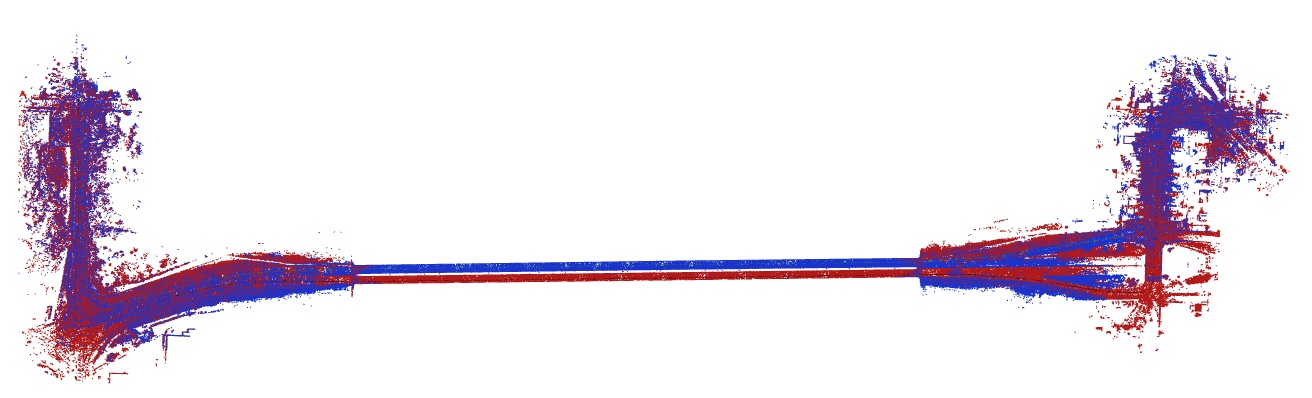}
    \caption{Two overlaid lidar maps of the \texttt{tunnel} route, aligned using ground-truth poses, constructed from sequences driven in opposite directions (red and blue).}
    \label{fig:tunnel_map}
\end{figure}

\begin{figure*}[t]
    \centering
    \includegraphics[width=\linewidth]{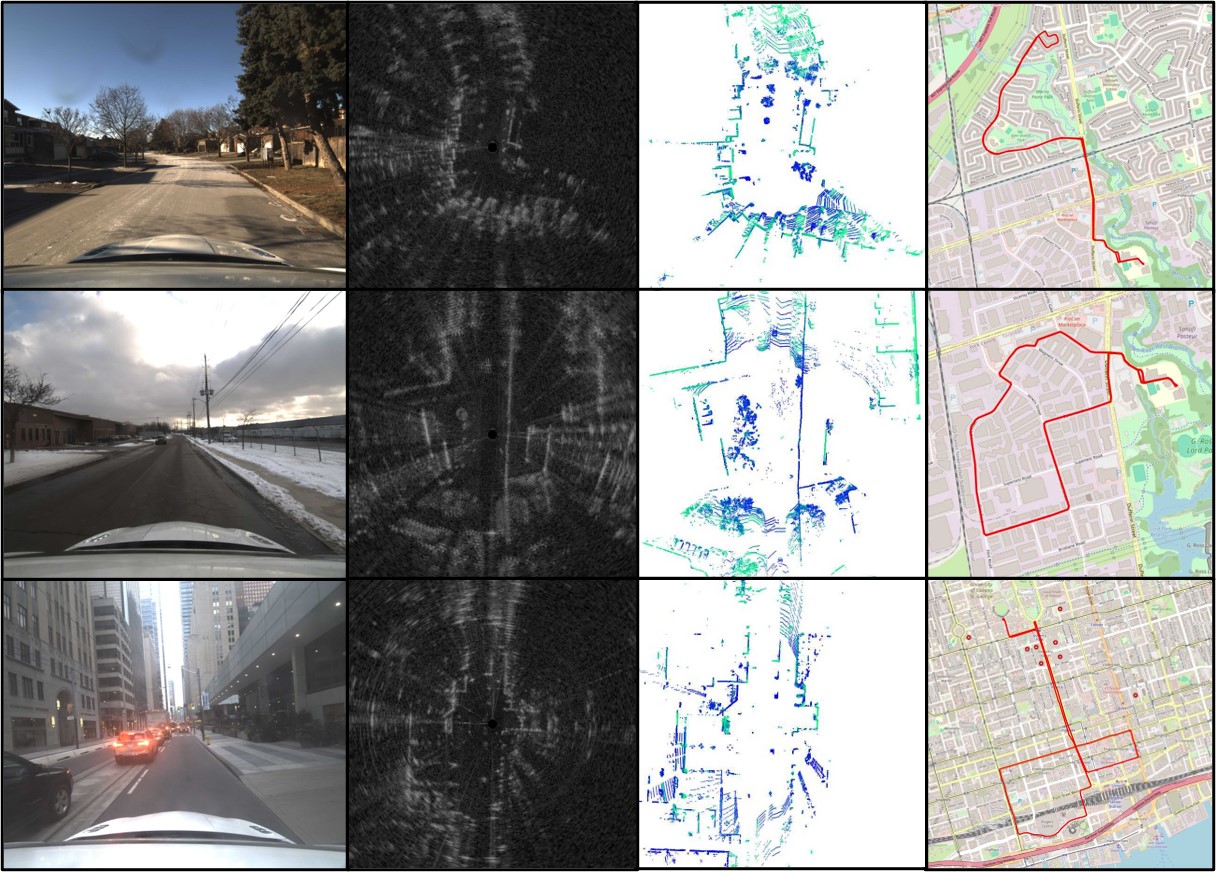}
    \caption{The `structured' routes. Top to bottom sequences: \texttt{suburbs}, \texttt{industrial}, \texttt{urban}. Left to right: camera image, radar scan, lidar scan (aligned to radar scan and with ground plane removed), OpenStreetMap route overview. Note the high level of structure and clear geometric features. The \texttt{urban} route features extensive cars, pedestrians, cyclists, trams, and other dynamic urban objects.}
    \label{fig:overlay_structured}
\end{figure*}

\begin{figure*}[t]
    \centering
    \includegraphics[width=\linewidth]{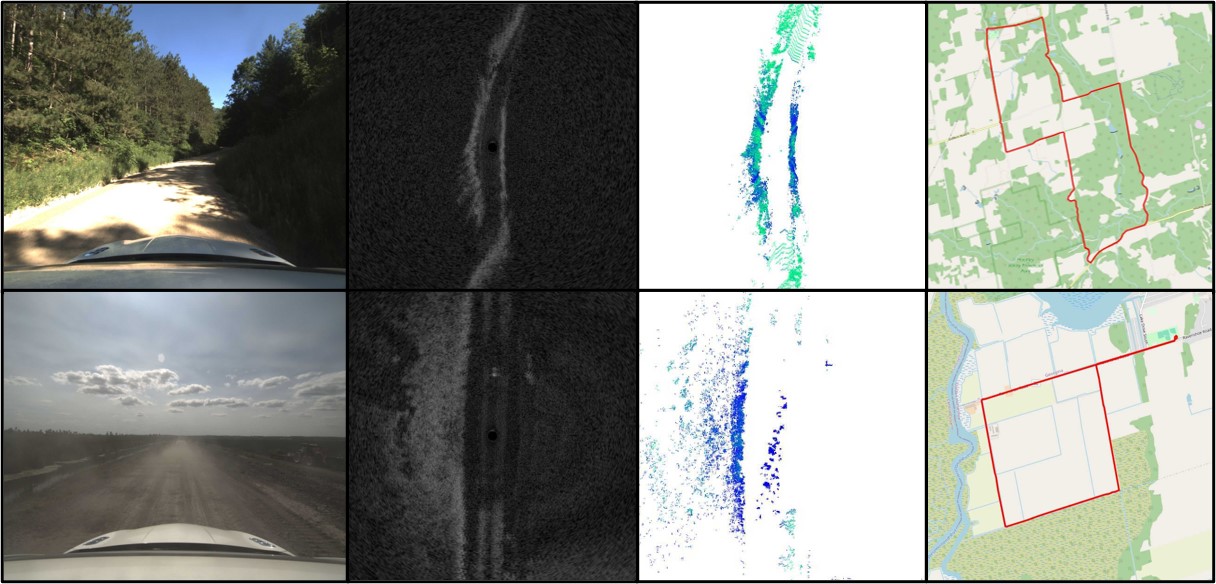}
    \caption{The `rural' routes. Top to bottom sequences: \texttt{forest}, \texttt{farm}. Left to right: camera image, radar scan, lidar scan (aligned to radar scan and with ground plane removed), OpenStreetMap route overview. Note the lack of any structure. Some \texttt{farm} sequences had a car drive in front of the data collection platform to raise a dust cloud, obscuring part of the camera and lidar returns.}
    \label{fig:overlay_rural}
\end{figure*}

\begin{figure*}[t]
    \centering
    \includegraphics[width=\linewidth]{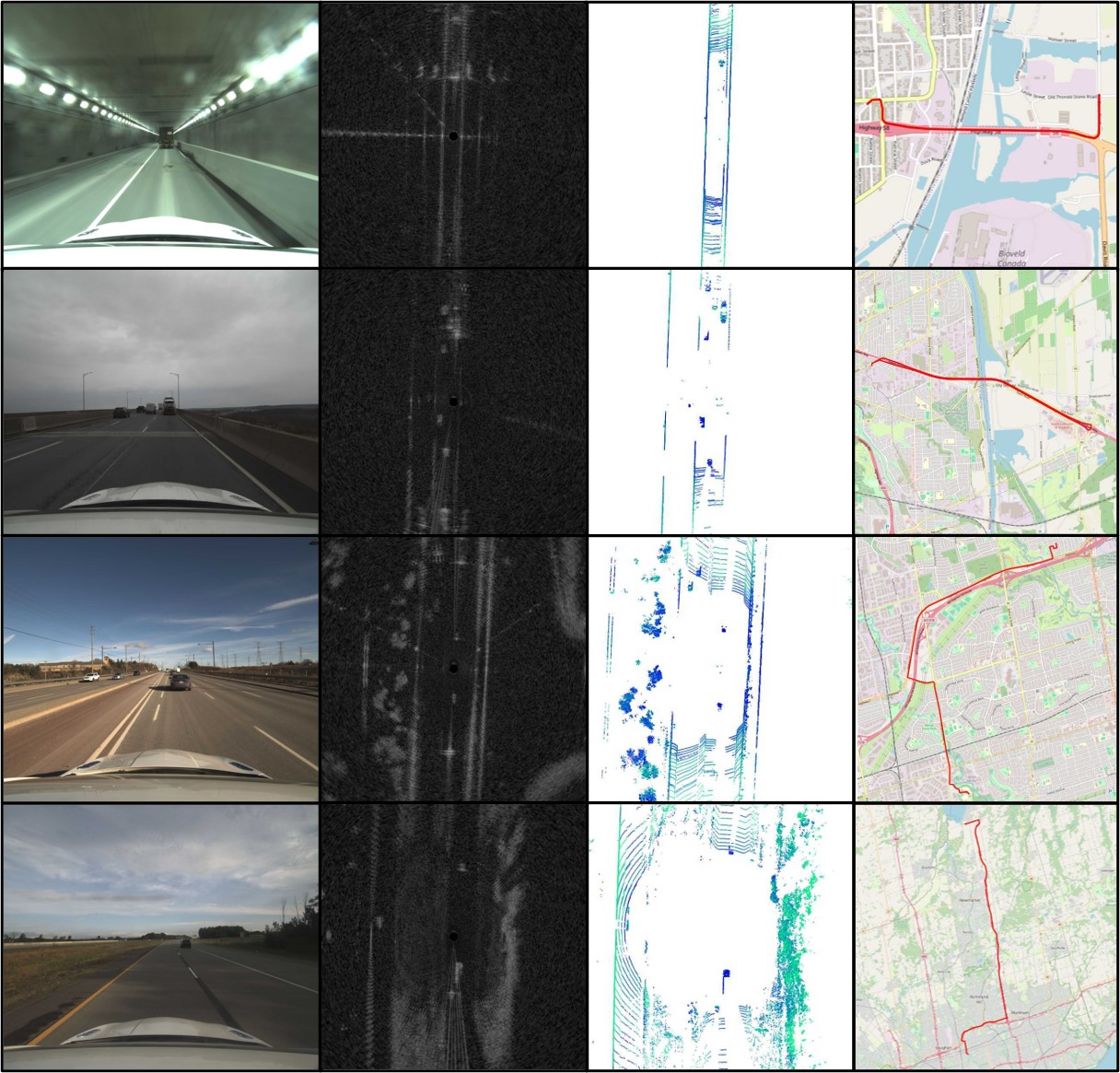}
    \caption{The `highway' routes. Top to bottom sequences: \texttt{tunnel}, \texttt{skyway}, \texttt{regional}, \texttt{freeway}. Left to right: camera image, radar scan, lidar scan (aligned to radar scan and with ground plane removed), OpenStreetMap route overview. Note the repetitive features parallel to the data collection vehicle and the stand-out vehicles as the main features.}
    \label{fig:overlay_highway}
\end{figure*}

\section{Sequences}

The dataset is composed of multiple traversals of nine different \textit{routes} covering a variety of urban, suburban, rural, and highway roads.
Each traversal, called a \textit{sequence}, covers anywhere from $1.9$ to $58\;\si{\km}$, for a total of $642.9\;\si{\km}$.
\rev{Sequences were collected over the course of 10 months from December 2024 to September 2025.}
Notable environmental conditions captured in each route are listed in Table~\ref{tab:sequences}.

The routes are grouped into three subsets: `structured', `rural', and `highway'.
Sequences are categorized as `loop' or `one-way'.
Loop sequences start and end at the same physical location for a given route.
One-way sequences alternate start and end locations, traversing the same roads in opposite directions; each direction is denoted by the nearest cardinal direction on the download page.
This enables evaluation of mapping in one direction and localization in the reverse.
On some routes, particularly the \texttt{tunnel}, sequences in opposite directions lose direct line-of-sight to each other, creating a rare yet realistic challenge.
Figure~\ref{fig:tunnel_map} illustrates this with two aligned lidar maps constructed from ground truth poses: one traversing the tunnel east (red) and the other west (blue).
The dataset is summarized in Table~\ref{tab:sequences}.

\subsection{Structured routes}
The `structured' subset, composed of the \texttt{suburbs}, \texttt{industrial}, and \texttt{urban} routes, is captured in environments that are densely developed.
This subset features constantly visible buildings, road signs/markings, and other general manmade objects.
Traffic conditions range from lightly busy to standstill traffic (in the \texttt{urban} route in particular) and surrounding dynamic `actors' are made up of cars of various sizes, semi-trucks, buses, trams, cyclists, and pedestrians.
Speed limits generally do not go above $60 \; \si{\km/\hour}$.
Visual examples of data collected during these routes, as well as the route shapes, are shown in Figure~\ref{fig:overlay_structured}.

\subsection{Rural routes}
The `rural' subset, composed of the \texttt{forest} and \texttt{farm} routes, is captured in environments that are almost entirely undeveloped.
Large portions of each route traverse dirt or gravel roads surrounded primarily by vegetation.
Speed limits go up to $80 \; \si{\km/\hour}$.

The \texttt{farm} sequences were collected over two days separated by one month, during which vegetation and roadside farming equipment changed substantially.
Because these elements constitute a large portion of the available scene information, such changes pose a significant challenge for localization.
Additionally, half of the \texttt{farm} sequences were driven while deliberately tailing another vehicle, which generated dust clouds on dirt road segments, making detection and localization tasks further challenging.
Visual examples of data collected during these routes, as well as the route shapes, are shown in Figure~\ref{fig:overlay_rural}.

\subsection{Highway routes}
The `highway' subset, composed of the \texttt{tunnel}, \texttt{skyway}, \texttt{regional}, and \texttt{freeway} routes, is captured in different kinds of highway environments.
This subset is characterized by long stretches of road with repetitive geometric features primarily alongside the vehicle, with this repetitiveness largely broken up only by other moving vehicles.
Speed limits go up to $110 \; \si{\km/\hour}$.
The \texttt{freeway} route is one of the more unique contributions of this dataset as it features over $50 \; \si{\km}$ of continuous freeway driving at high speed.
Visual examples of data collected during these routes, as well as the route shapes, are shown in Figure~\ref{fig:overlay_highway}.

\section{Data format}

\subsection{Data organization}
\begin{figure}
    \centering
    \includegraphics[width=0.58\linewidth]{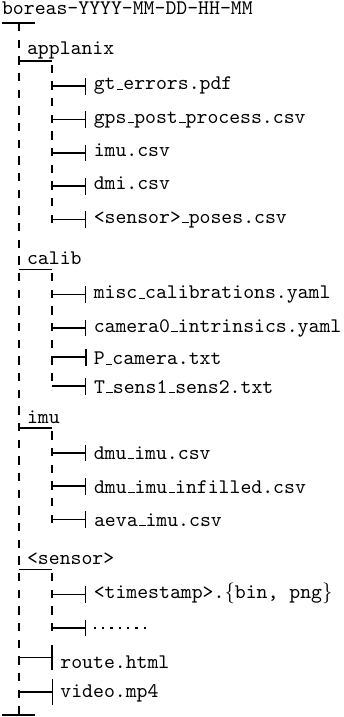}
    \caption{The data organization structure for a Boreas-RT sequence. Each \texttt{<sensor>} directory is repeated for \texttt{lidar} (Velodyne), \texttt{radar}, \texttt{camera}, and, when available, \texttt{aeva}. The \texttt{imu.csv} file contains Applanix IMU data used to create the ground truth. The \texttt{aeva\_imu.csv} file is present only in sequences containing Aeva data.}
    \label{fig:data_org}
\end{figure}

Following the same organizational structure as the original Boreas dataset, Boreas-RT is split into \textit{sequences} that include all data and ground truth from a single recording.
The sequences also include a calibration folder, making them entirely self-contained for running experiments.
Sequences are named based on the timestamp at which data collection was started following the format \texttt{boreas-YYYY-MM-DD-HH-MM}.
Figure~\ref{fig:data_org} shows the file layout of each sequence.

\subsection{File formats}
All raw data filenames are given as UNIX epoch times in microseconds, with time synchronized to the Applanix POS LV UTC time.
The section on calibration provides details about the temporal calibration of all sensors.
Note that while all sensors are synchronized to the same clock, individual measurements are not synchronous with each other.
For example, this means that the $10\;\si{\hertz}$ Velodyne and $10\;\si{\hertz}$ Aeva data will have scans triggered at independent times and scan indices will not correspond to the same timestamp.
However, given the precise temporal synchronization and accurate relative pose information, measurement fusion can still be easily implemented.

\subsubsection{Camera:}
Camera data is saved as \texttt{<timestamp>.png} files, with $\mathrm{\texttt{<timestamp>}}  = t_\textrm{event} + t_\textrm{exposure}/2$, where $t_\textrm{event}$ is the timestamp at which an image capture is initiated, and $t_\textrm{exposure}$ is the total exposure time for the given image.
This sets the timestamp to the middle of the image capture event.

\subsubsection{Lidar:}
Lidar scans are saved each time the lidar completes a full 360\si{\degree} rotation.
All points collected during the rotation are compiled into a single \texttt{<timestamp>.bin} file, with \texttt{<timestamp>} set to the temporal middle of the scan.
Each point contains six fields $[x, y, z, i, d, t]$: the $(x, y, z)$ position of the point, the intensity $i$ of the return, the ID $d$ of the laser beam used to produce the point, and the global timestamp $t$ at which that point was recorded.

\subsubsection{Aeva:}
\begin{figure}[t]
    \centering
    \includegraphics[width=\linewidth]{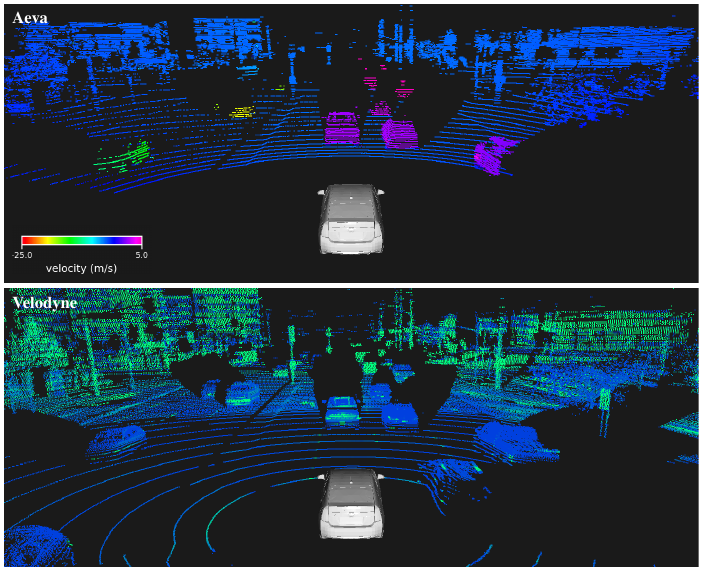}
    \caption{An Aeva and Velodyne scan captured at the same point in a sequence. The Aeva scan is coloured by per-point Doppler velocity, and the lidar scan by intensity.}
    \label{fig:aeva_example}
\end{figure}

Aeva scans are saved in the same fashion as Velodyne ones, with a single sweep of the entire field of view captured in a \texttt{<timestamp>.bin} file.
The \texttt{<timestamp>} is set to the temporal start of the scan.
Each point contains 9 fields $[x, y, z, v, i, q, r, t, f]$: the $(x, y, z)$ position of the point, the relative radial velocity measurement $v$ of the point, the intensity $i$ of the return, the signal quality $q$ of the return, the calibrated reflectivity $r$ of the detection, the time offset $t$ relative to the temporal start of the scan, and a 64-bit bitset $f$ encoding per-point flags.
Figure~\ref{fig:aeva_example} shows an example of Aeva data coloured by the per-point Doppler velocity estimates.

\subsubsection{Radar:}
\begin{figure*}
    \centering
    \includegraphics[width=0.98\textwidth]{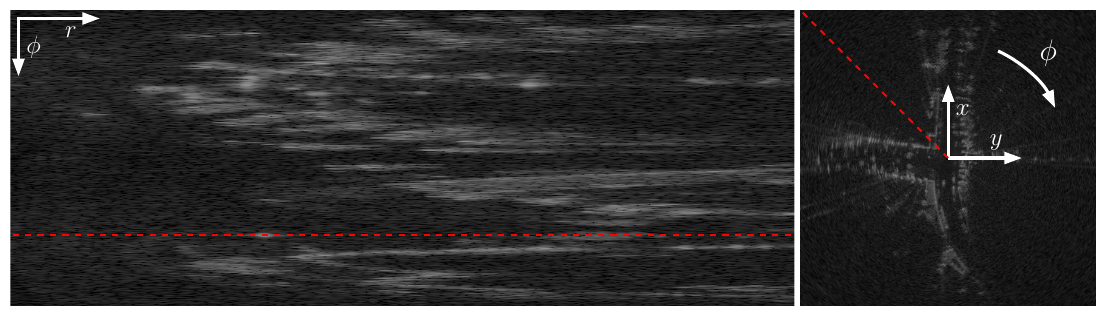}
    \caption{Left: An example raw polar radar image with metadata and much of the range trimmed off for clarity. The image shows intensity returns at different ranges $r>0$ for different azimuths $\phi \in [0\si{\degree},360\si{\degree})$. Right: The polar image unwrapped into radar-frame Cartesian coordinates $x$ and $y$. The dashed red line represents the same data in both images.}
    \label{fig:radar_data_example}
\end{figure*}

Radar scans, similar to lidar ones, correspond to data from a full 360\si{\degree} rotation of the sensor.
Each full rotation is captured in an $M \cross R$ 2D \texttt{<timestamp>.png} image, with $M = 400$ azimuths and $R=6859$ range and metadata bins.
Following the notation introduced by \cite{oxford_radar_robotcar} and used by \cite{burnett2023boreas}, the first eleven pixels (bytes) for each azimuth correspond to metadata associated with that azimuth: the first eight pixels contain the UNIX epoch time of the azimuth as a 64-bit integer, then two pixels contain the rotational encoder value as a 16-bit unsigned integer, and finally, the last column contains the `chirp type' of the azimuth that communicates whether frequency was modulated up or down for the given azimuth.
This chirp type facilitates the extraction of Doppler velocity measurements from the radar data (see \cite{lisus2025doppler} for an example).
It is additionally required in order to implement Doppler undistortion of the data \citep{2021_Burnett}, since `up-chirps' have a positive Doppler distortion effect and `down-chirps' have a negative Doppler distortion.
Figure~\ref{fig:radar_data_example} shows an example of a cropped radar image and the same image `unwrapped' into radar frame Cartesian coordinates.
Our development kit (devkit) provides an implementation of the `unwrapping' function.

\subsubsection{IMU:}
All IMU data is contained in \texttt{.csv} files with a header explaining what each column contains.
The biggest difference is in the units of the timestamps associated with each IMU measurement: the DMU41 IMU reports time in nanoseconds, the Aeva IMU reports time in microseconds, and the Applanix IMU reports time in seconds.
At a minimum, each IMU contains angular velocity $w_x, w_y, w_z$ and linear acceleration $a_x, a_y, a_z$ measurements at timestamps $t$ resolved in the respective IMU frame.

\subsubsection{Wheel encoder:}
The wheel encoder data is contained in a distance measurement indicated (DMI) file \texttt{dmi.csv}.
This file contains only two columns: \texttt{GPSTime} in seconds and \texttt{pulse\_count} denoting the number of pulses detected by the wheel encoder.
\begin{figure}[t]
    \centering
    \includegraphics[width=\linewidth]{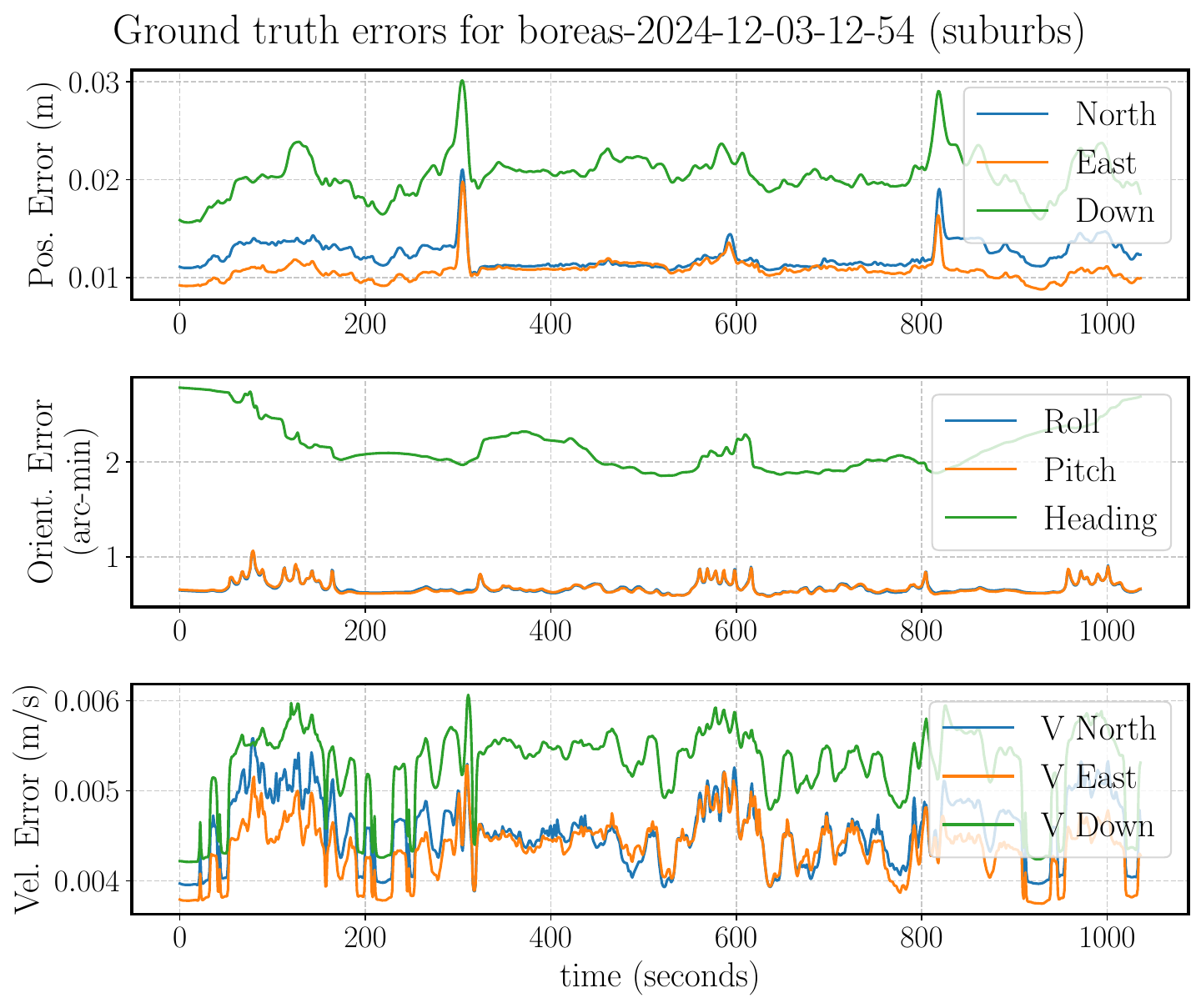}
    \includegraphics[width=\linewidth]{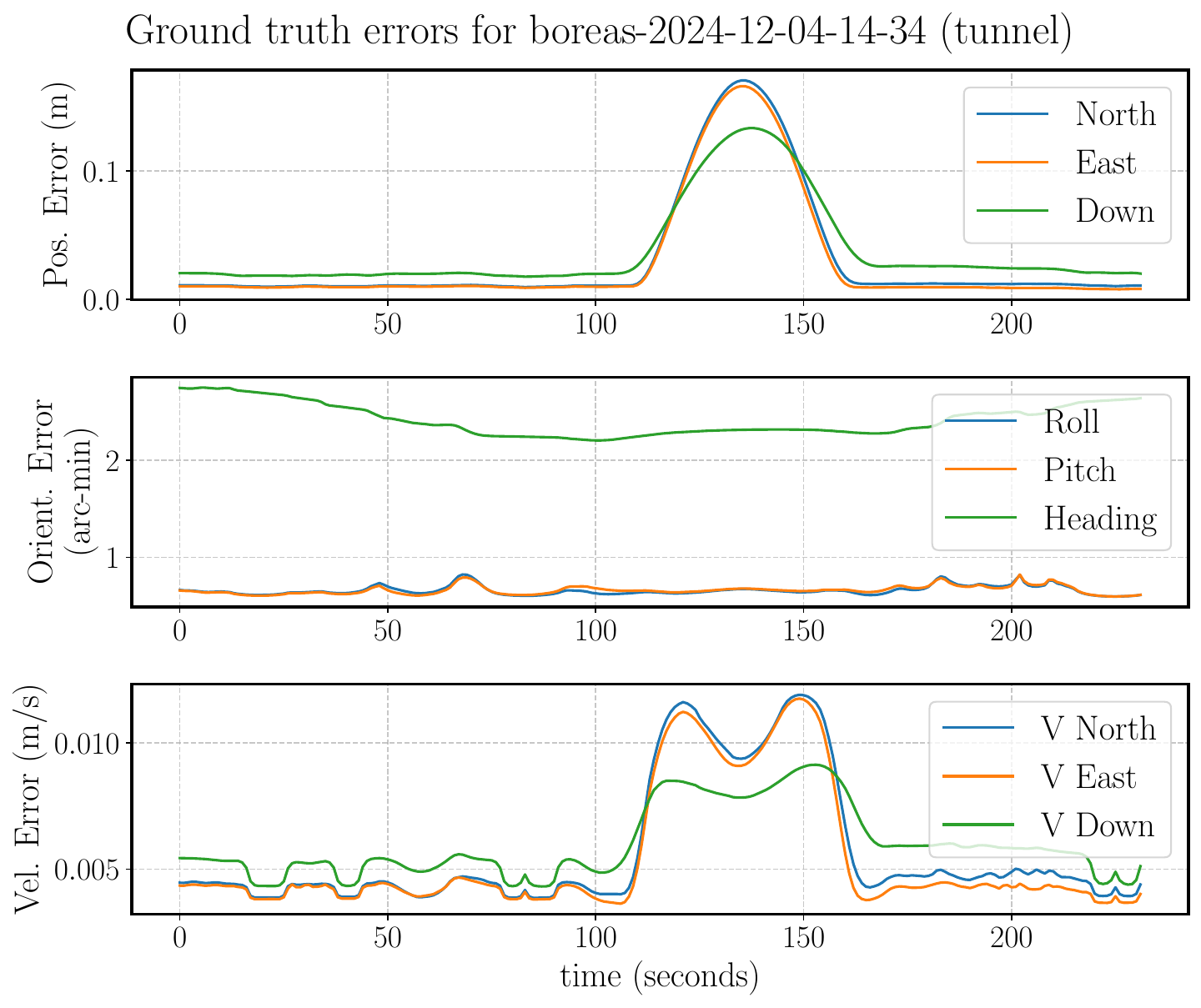}
    \caption{Self-reported ground truth errors from the Applanix POSPac software. Each set of three plots shows the position errors, orientation errors, and linear velocity errors for each axis. Top: Ground truth errors for a suburban sequence with good GPS visibility throughout the sequence. Bottom: Ground truth errors for a tunnel sequence where GPS connection was lost while inside the tunnel from around 110 to 160 \si{\second}.}
    \label{fig:gt_errors}
\end{figure}

\section{Ground truth}
Ground truth is generated using a similar approach as in the original Boreas dataset: GPS, IMU, and wheel encoder measurements are post-processed using Applanix's POSPac software suite with RTK-based corrections.
\rev{This software runs a batch optimization over the entirety of each sequence using all of these sources of information to compute the pose and velocity of the Applanix reference frame.}
The post-processed ground truth is reported at $200\;\si{\hertz}$ in the \texttt{applanix/gps\_post\_process.csv} file.
Poses are reported in East-North-Up (ENU) coordinates, with $x$-$y$-$z$ corresponding to the respective directions.
For users to assess the ground truth quality for each sequence, a visualization of the position, orientation, and linear velocity errors is provided in the \texttt{applanix/gt\_errors.pdf} file.
Two examples of such visualizations are shown in Figure~\ref{fig:gt_errors}.
The top image shows errors in nominal conditions, where the platform had an unobstructed view of the sky throughout the data collection period and operated relatively close to a base station, enabling optimal RTK corrections.
Many of the sequences have errors of a similar order of magnitude, with the East and North position errors remaining below $2\;\si{\cm}$ and the up errors remaining below $3\;\si{\cm}$ throughout the entire sequence.
This is an improvement over the original Boreas dataset, in which the East and North errors were in the $2$--$5\;\si{\cm}$ range and the up errors on the order of $5$--$10\;\si{\cm}$.
A similar improvement can be observed in the orientation and velocity errors and is a result of switching from RTX to RTK for post-processing.
These errors show the global accuracy of the ground truth, meaning that relative frame-to-frame errors are expected to be even lower.

\rev{The bottom of Figure~\ref{fig:gt_errors} shows errors in more challenging conditions, where the platform enters a tunnel for approximately $50\;\si{\second}$.
Throughout this period, no satellite connection could be established and the position and linear velocity errors increased.
Challenging conditions occasionally yield position errors on the order of $10$--$80\;\si{\cm}$ owing to tunnels, extreme urban canyon effects, or remote locations far from base stations.
These segments, especially if they occupy a significant portion of the route, can make localization errors dominated by ground truth inconsistency.
The \texttt{urban} route, owing to the urban canyon effect on GPS, is particularly affected, with ground truth translational component differences of up to $80\;\si{\cm}$ at the same location between sequences.
We omit this route from the localization results, and suggest others do the same, on account of expected localization performance to be far below this level.
The \texttt{forest} route, due to being far away from the nearest base station, is affected to a lesser extent with ground truth translational differences of up to $40\;\si{\cm}$.
We keep this route in the localization results on account of it being challenging to successfully localize within to any degree of accuracy.
Finally, the \texttt{tunnel} route can have errors up to $15\;\si{\cm}$ during the tunnel portions, although we again keep it due to the difficulty of achieving any successful localization in such a geometrically degenerate environment.
We provide an approximate upper bound on the expected per-axis global accuracy for each route in Table~\ref{tab:sequences}, which can be interpreted as the best possible localization accuracy.
Algorithms capable of localizing to below this bound will not necessarily show improved performance, since the error computation will become dominated by ground truth noise.
We note that local accuracy, which is required to evaluate odometry, is still expected to be at a centimetre level for the \texttt{forest}, \texttt{urban}, and \texttt{tunnel} routes despite their decimetre-level global inconsistency.}

For convenience, the GPS ground truth is interpolated at each sensor's timestamps, transformed into each sensor's frame, and stored into individual \texttt{applanix/<sensor>\_poses.csv} files.
Each line of a sensor's ground truth reports data at the epoch timestamp in microseconds $t$ corresponding to the timestamp encoded in the measurement's name.
Each line provides the position of the sensor $s$ with respect to a fixed point $e$ in the ENU frame $\mathcal{F}_e$ as measured in the ENU frame $\mbf{r}_e^{se} = \begin{bmatrix}x & y & z \end{bmatrix}^\trans$ and the roll $r$, pitch $p$, and yaw $y$ (heading) angles that can be used to find the rotation from the sensor frame $\mathcal{F}_s$ to the ENU frame $\mbf{C}_{es} = \mbf{C}_1(r)\mbf{C}_2(p)\mbf{C}_3(y) \in SO(3)$ \citep{barfoot2024state}.
The full 3D pose at each timestamp can be formed as
\begin{align}
    \mbf{T}_{es} = \begin{bmatrix}
        \mbf{C}_{es} & \mbf{r}_e^{se} \\ \mbf{0}^\trans & 1
    \end{bmatrix} \in SE(3).
\end{align}
Each line also reports the linear velocity of the sensor with respect to the ENU frame $\mbf{v}_e^{se}~=~\begin{bmatrix}v_x & v_y & v_z \end{bmatrix}^\trans$ and the angular velocity of the sensor with respect to the ENU frame as measured in the sensor frame $\mbs{\omega}_s^{se}~=~\begin{bmatrix}\omega_x & \omega_y & \omega_z \end{bmatrix}^\trans$.
All of these quantities are formatted as $[t, x, y, z, v_x, v_y, v_z, r, p, y, \omega_z, \omega_y, \omega_x]$ (note the flipped ordering of the angular velocities for backward compatibility).
\section{Calibration}

\subsection{Temporal calibration}
All sensors were temporally synchronized to UTC time as reported by the Applanix POS LV.
The data-recording computer and Velodyne lidar were synchronized to the Applanix clock through a hardwired connection carrying a PPS signal and NMEA time messages.
The camera was set to output a square-wave pulse whose rising edge marked the start of each exposure, and the Applanix POS LV was configured to detect and timestamp these pulses.
The Navtech radar and Aeva lidar were synchronized to the data-recording computer via PTP; because the computer itself was synchronized to the Applanix system, both sensors were therefore aligned to the Applanix clock.

The stand-alone DMU41 IMU did not have the capability to be directly synchronized to the Applanix clock.
Instead, each sequence started and ended with a `jerk' of the car.
The forward acceleration recorded during the jerk periods by the DMU41 IMU and the internal Applanix IMU were compared and aligned using cross-correlation. 
The temporal offset estimated by the alignment was then averaged between the start and end jerk periods.
The data used for this temporal alignment, and all data that was collected during the jerks, was discarded to prevent correlations between IMU measurements.
\rev{The IMU timestamps were also found to drift slightly and would occasionally have dropouts due to a bug in our driver implementation.
The DMU41 operates at a hardware-specified rate of $200.2 \, \mathrm{Hz}$, confirmed empirically, rather than the nominal $200.0 \, \mathrm{Hz}$.
Each raw IMU message contained a drift-affected timestamp $t$ and a clock-independent message count $\mathrm{count}$, which allowed us to recompute $\mathrm{t} = \mathrm{t_{start}} + \mathrm{count} / \mathrm{200.2}$, where $\mathrm{t_{start}}$ is the sequence start time aligned to the Applanix clock using the initial jerk.
This approach was verified by using the jerk-based, Applanix-aligned sequence end timestamp $\mathrm{t_{end}}$ and confirming that $\mathrm{t_{end}} = \mathrm{t_{start}} + \mathrm{count_{end}} / \mathrm{200.2}$ to within the sensor-declared accuracy of $0.5 \, \mathrm{ms}$.}
Though rare, the IMU dropout periods can be as long as $0.4\;\si{\second}$, meaning that multiple scans/images of exteroceptive sensor data could occur without any IMU measurements.
This provides researchers with an opportunity to implement robust algorithms using real data, as data dropouts are possible for a variety of reasons.
To alleviate the dropout issue and allow for testing of algorithms with `ideal' data, we also provide an `infilled' \texttt{dmu\_imu\_infilled.csv} IMU data file, which resamples the raw sensor data at an exact $200\;\si{\hertz}$ using simple linear interpolation.

\subsection{Intrinsic calibration}
\subsubsection{Camera:}
Camera intrinsics and distortion parameters were calibrated using MATLAB’s Camera Calibrator \citep{mathworks_camera_calibrator_app} and are stored in \texttt{camera0$\_$intrinsics.yaml}.
To estimate these parameters, images of a calibration checkerboard were captured at various distances and orientations.
From this static data collection, a subset of images was selected based on checkerboard pose diversity and the calibration solutions they produced, retaining images that resulted in a model with consistently low reprojection error.
The computed intrinsics were used to undistort the raw images and to compute an optimal new camera matrix $\mbf{P}$ of the form
\begin{align}
    \mbf{P} = \begin{bmatrix}
        f_u & 0 & c_u & 0 \\
        0 & f_v & c_v & 0 \\
        0 & 0 & 1 & 0 \\
        0 & 0 & 0 & 1
    \end{bmatrix},
\end{align}
where $f_u$, $f_v$ are the focal lengths and $c_u$, $c_v$ are the principal offsets for image axes $u$ and $v$.
This matrix is provided under \texttt{P$\_$camera.txt}.
The dataset provides already radially undistorted images corresponding to $\mbf{P}$.

\subsubsection{Radar:}
When extracting range information from the radar, we must account for a range offset, which needs to be subtracted when converting pixels to range values.
This offset occurs because the radar signal needs to travel some distance within the sensor before being emitted.
The RAS6 manufacturer Navtech reports this range offset as $-0.31\;\si{\meter}$ and we report this value in the \texttt{calib/misc\_calibrations.yaml} file.
\rev{Note that this offset has not been incorporated into the radar data.}
We also include the hardware-dependent $\beta$ parameter (\cite{lisus2025doppler}) relating the Doppler velocity to the range offset it produces in \texttt{calib/misc\_calibrations.yaml}

\subsubsection{Velodyne:}
The only notable `intrinsic' for the Velodyne lidar is the presence of a faulty beam (beam ID 69).
This beam consistently under-reports the range of points and was consequently removed from the saved pointclouds.
This fault is likely due to sensor degradation over the years, as it was not noted in the original Boreas dataset.

\subsubsection{Wheel encoder:}
The wheel radius is calibrated by comparing the measured ground truth velocity with the velocity inferred from wheel encoder pulses, which are converted to speed by multiplying the distance travelled per pulse by the number of pulses per second (the pulse rate).
An initial estimate of the distance travelled per pulse is computed from the average ratio of ground truth velocities to pulse rate.
A least-squares optimization adjusts the distance per pulse to minimize the difference between the ground truth and inferred velocities, accounting for the angular velocity and the lever arm between the Applanix and the wheel.
The estimated distance per pulse is then converted to the wheel radius by multiplying by the encoder resolution and dividing by $2\pi$.
The calibrated radius is reported in the \texttt{calib/misc\_calibrations.yaml} file.

\begin{figure}[t]
    \includegraphics[width=0.49\textwidth]{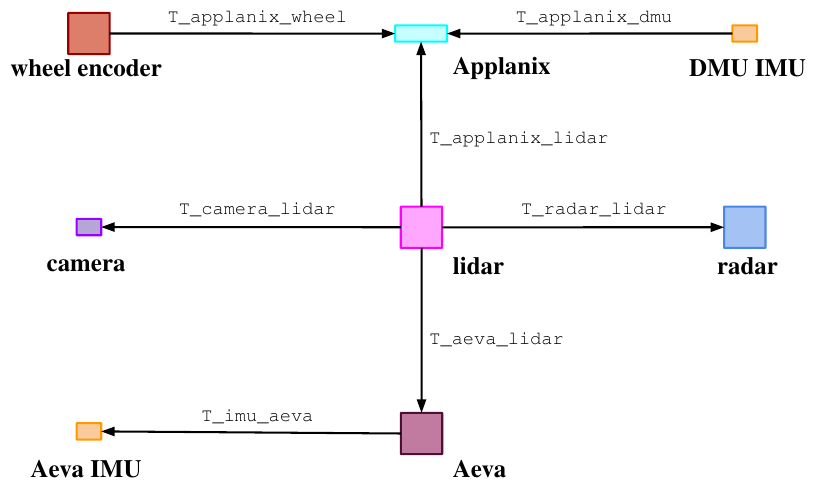}
    \caption{Transform tree for Boreas-RT extrinsic calibrations. All exteroceptive sensors are calibrated relative to the Velodyne lidar, which provides rich 360\si{\degree} environmental coverage. The specific calibration file names under \texttt{/calib} are shown.}
    \label{fig:tf_tree}
\end{figure}

\subsection{Extrinsic calibration}
The full extrinsic transform tree is visualized in Figure~\ref{fig:tf_tree}.
Exteroceptive sensors are calibrated relative to the rich 360\si{\degree} Velodyne lidar, with the Velodyne calibrated against the Applanix-provided ground truth.
Extrinsics were calibrated either using static data or by using two hold-out \texttt{suburbs} sequences that `bookend' the dataset: one was collected just prior to the first sequence and one just after the last sequence included in Boreas-RT.
All extrinsic transformations are reported as $\mbf{T}_{ab} \in SE(3)$ matrices that define the transformation from frame $\mbs{\mathcal{F}}_b$ to frame $\mbs{\mathcal{F}}_a$ \citep{barfoot2024state}.
The naming scheme of the transformation files follows the same convention.

\subsubsection{Velodyne to Applanix:}
A lidar odometry algorithm \citep{are_we_ready_for} was run on two hold-out \texttt{suburbs} sequences to align the Velodyne and Applanix frames by matching the estimated linear velocity to the Applanix ground-truth velocity.
Only yaw was optimized as translation, roll, and pitch are weakly observable under largely planar vehicle motion.
Translation was taken from the sensor-rack CAD model \citep{burnett2023boreas}, with a non-zero $z$ component due to vertical stacking, and roll and pitch were set to zero.

\subsubsection{Velodyne to camera:}
\begin{figure}[t]
    \centering
    \includegraphics[width=0.5\linewidth]{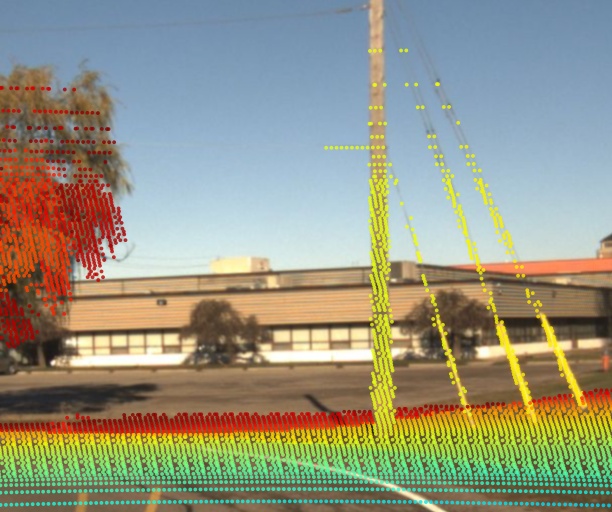}%
    \includegraphics[width=0.5\linewidth]{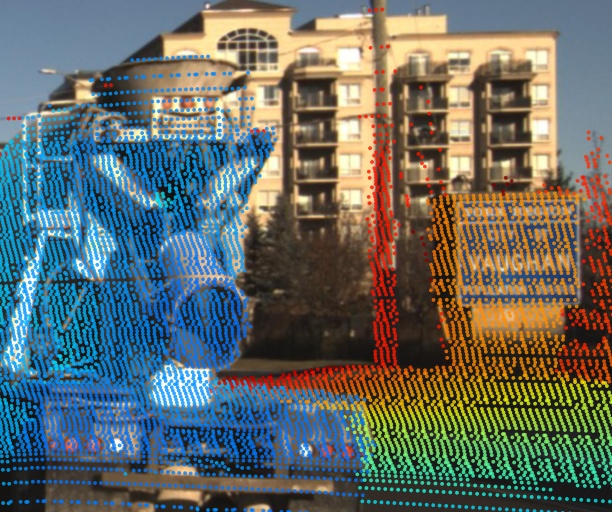}

    \vspace{-0.2em}

    \includegraphics[width=0.5\linewidth]{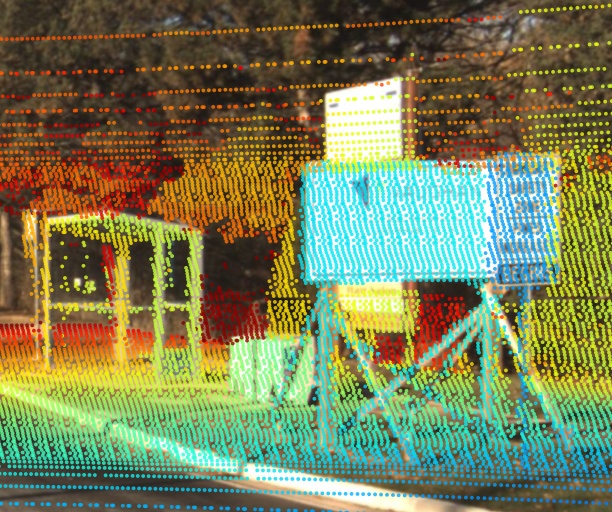}%
    \includegraphics[width=0.5\linewidth]{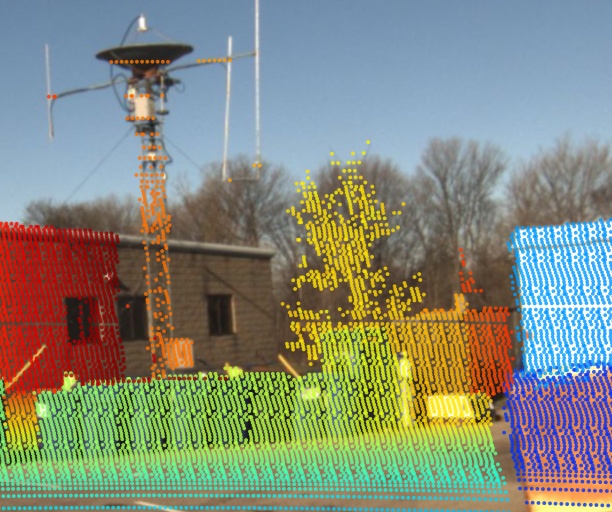}

    \caption{\rev{Zoomed-in camera--Velodyne alignment across four representative samples.}}
    \label{fig:cam_lidar_calib}
\end{figure}

The Velodyne was calibrated to the camera using MATLAB’s Camera to Lidar Calibrator \citep{mathworks_lidar_camera_calibrator_app}.
Static data with a checkerboard in different positions and orientations in the scene, different from the data used for the camera intrinsic calibration, was used for the calibration.
Alignment was achieved by minimizing the reprojection, translation, and rotation errors, which quantify pixel alignment and geometric consistency of the checkerboard plane between the two modalities.
Figure~\ref{fig:cam_lidar_calib} projects Velodyne points onto a camera image to demonstrate the quality of the calibration.







\subsubsection{Velodyne to radar:}
\begin{figure}[t]
    \centering
    \includegraphics[width=0.65\linewidth, trim={2cm 3cm 2cm 3cm},clip]{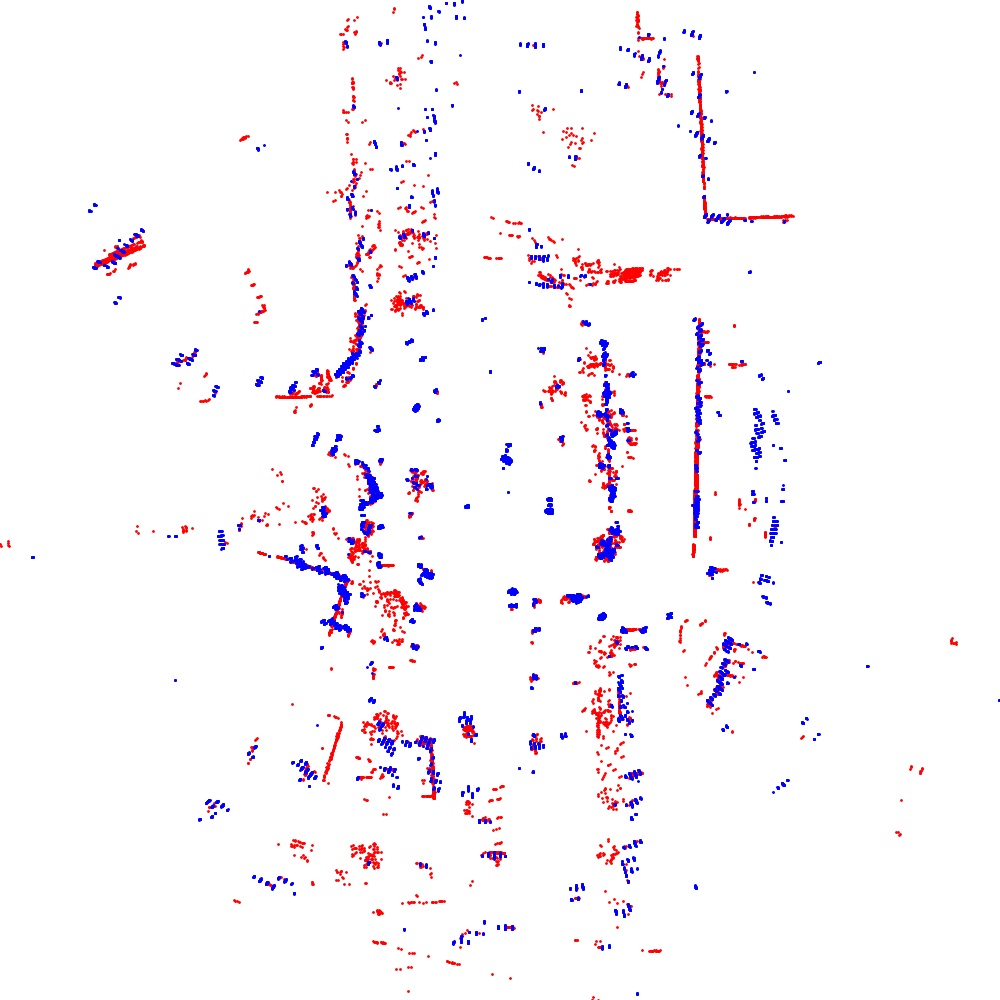}
    \caption{A bird's-eye-view of radar points (blue), extracted using the $K$-strongest pointcloud extractor, overlaid with Velodyne lidar points (red) projected into the radar frame using the calibrated Velodyne to radar extrinsics.}
    \label{fig:radar_lidar_calib}
\end{figure}

Velodyne to radar extrinsic calibration was performed using a 2D Gauss–Newton iterative closest point (ICP) optimization applied jointly over multiple scans to estimate the yaw offset between the sensors.
Calibration data consisted of radar–Velodyne scan pairs collected at 25 stationary poses, with 5 scans per pose, gathered across two sessions: one just prior to the first sequence and one just after the last sequence released for Boreas-RT.
For each scan, a radar pointcloud was extracted using the $K$-Strongest extractor \citep{finer_points} and a corresponding 2D Velodyne pointcloud was obtained by selecting points with reliable normal scores and projecting those within the radar’s field of view onto a 2D plane.
Correspondences were formed and residuals evaluated across all scans with one shared transform, which was iteratively updated until convergence.
A Cauchy loss was used to reduce the influence of spurious matches on the optimization.
The translation components were taken directly from the CAD model of the sensor rack \citep{burnett2023boreas}, with only the $z$ component being non-zero due to the vertical stacking of the radar and the Velodyne.
The roll and pitch were set to 0 since they are unobservable from the 2D radar data.
Figure~\ref{fig:radar_lidar_calib} demonstrates the quality of the calibration.

\subsubsection{Velodyne to Aeva:}
\begin{figure}[t]
    \centering
    \includegraphics[width=1.0\linewidth, trim={7cm 5cm 7cm 6cm},clip]{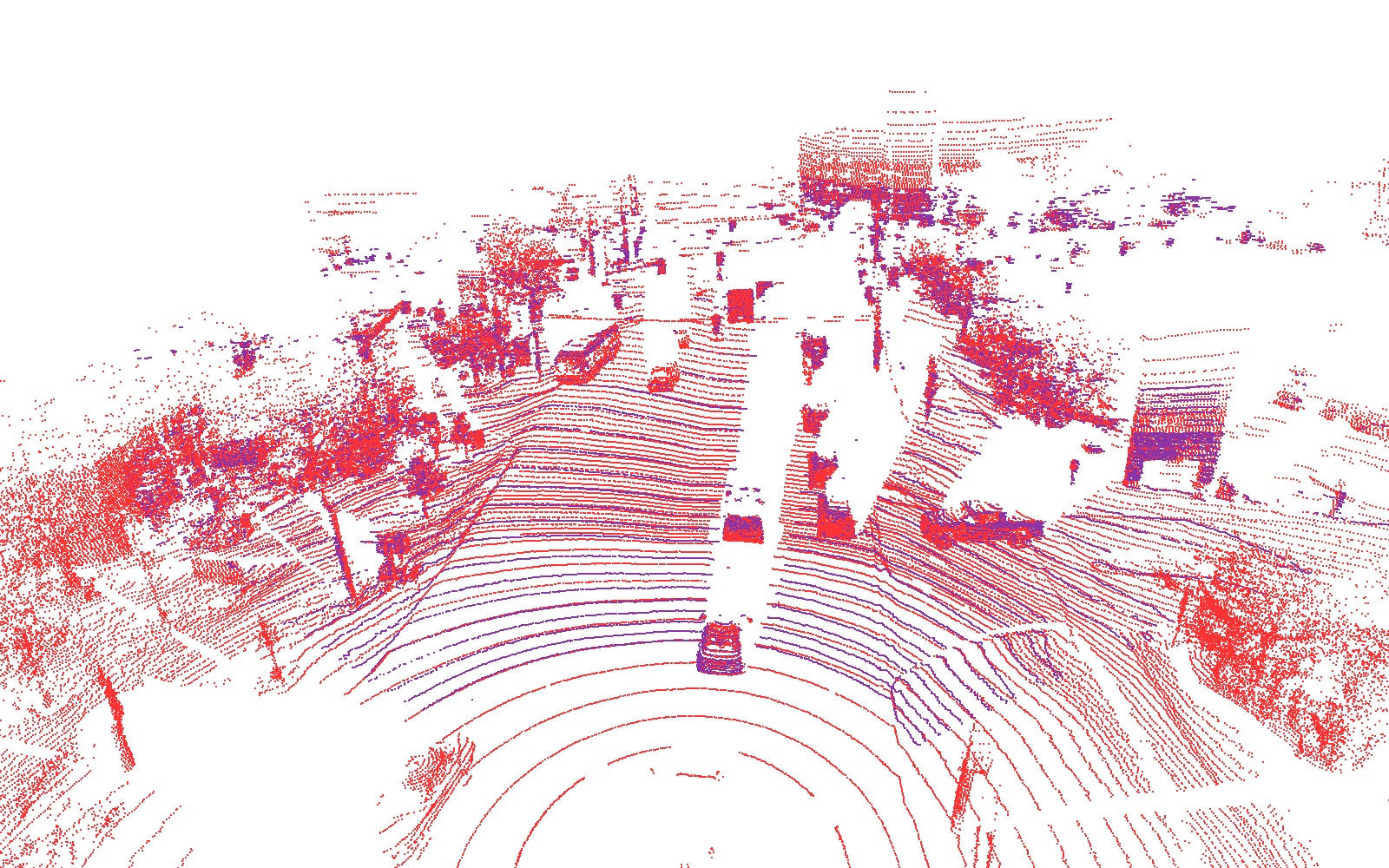}
    \caption{Aeva points (purple) and Velodyne points (red) transformed into the Aeva frame using the calibrated Velodyne to Aeva extrinsics.}
    \label{fig:aeva_lidar_calib}
\end{figure}

The Velodyne to Aeva extrinsic calibration was obtained using a 3D point-to-plane ICP formulation applied across multiple scans, with Gauss–Newton optimization and a Cauchy robust loss used to achieve stable alignment.
The calibration data consisted of 140 Aeva-Velodyne pointcloud pairs collected over 20 stationary poses prior to the first Boreas-RT sequence collection.
As in the Velodyne–radar calibration, a single extrinsic transform was jointly estimated across all scan pairs, with correspondences and residuals evaluated over all scans in the set.
Full 3D calibration was possible on account of both sensors containing dense 3D data.
Figure~\ref{fig:aeva_lidar_calib} shows a visual alignment of the two sensors using the optimized extrinsics.

\subsubsection{Aeva to Aeva IMU:}
The Aeva lidar contains a built-in IMU, the data from which we include in all sequences containing Aeva data.
There is a small translational offset between the lidar and the IMU reported by the manufacturer, which is captured in the extrinsic transform file \texttt{T$\_$imu$\_$aeva.txt}.
The Aeva IMU is axis- and time-aligned with the Aeva lidar.

\subsubsection{Wheel encoder to Applanix:}
The wheel encoder to Applanix transformation was obtained by projecting the wheel encoder velocities into the Applanix frame using the current extrinsic estimate, and comparing them to the measured ground truth velocities.
The optimization adjusted the rotation and lever arm of the transform to minimize the residual velocity error.
This process was iteratively repeated until the projected wheel velocities best aligned with the ground truth across the hold-out sequences.

\subsubsection{IMU to Applanix:}
The DMU41 IMU was calibrated to the Applanix by solving Wahba’s problem \citep{Wahba_1965} on angular velocity measurements from both sensors over the two hold-out sequences to estimate the full 3D orientation.
Because translation is weakly observable, the $x$ and $y$ offsets were set to zero, and the vertical $z$ offset was measured by hand, consistent with the sensors’ vertical stacking.

\section{Benchmarks}
\label{sec:benchmarks}

To highlight the value of this dataset, we evaluate it using SOTA odometry and localization algorithms.
Results are shown in $SE(2)$ for algorithms that are capable of only 2D estimation and in $SE(3)$ for those that can generate full 3D results.
These benchmarks show that existing approaches, despite SOTA performance in their original publications, do not maintain the same performance and may even fail on road types not considered during development.
To this end, we use parameters tuned to the data for which the algorithms were primarily developed: almost exclusively \texttt{suburbs} and \texttt{industrial}-style routes.
We argue that a SOTA algorithm should perform with the same level of accuracy, using the same parameter set, in all road types and conditions.

\subsection{Odometry}
Benchmarked odometry algorithms are presented below, and each uses gyroscope data from the DMU41 IMU.
\begin{itemize}
    \item DRO \citep{legenti2025dro}: A direct radar odometry pipeline that leverages radar scans and a heading gyroscope to provide $SE(2)$ odometry estimates. DRO directly leverages the alternating-chirp-based Doppler information contained in the new radar scans.
    \item RTR \citep{are_we_ready_for}: The odometry branch of a radar teach \& repeat (RTR) implementation that extracts a pointcloud from radar scans using the $K$-strongest extractor and aligns consecutive pointclouds in a continuous-time (CT) manner using ICP. This implementation is supplemented with a gyroscope for best results \citep{lisus2025doppler}. RTR is only capable of producing $SE(2)$ estimates.
    \item LTR \citep{are_we_ready_for}: The odometry branch of a lidar teach \& repeat (LTR) implementation that aligns lidar pointclouds in CT using ICP, supplemented with gyroscope factors for best performance. LTR is tested using Velodyne data, although the same formulation can work on Aeva data \citep{balancing_act}. This method produces full $SE(3)$ state estimates, from which the $SE(2)$ components are extracted and evaluated separately for the $SE(2)$ benchmarks.
    \item D-Aeva \citep{need_for_speed}: A Doppler-inertial odometry estimator that leverages the per-point Doppler velocity measurements from Aeva and gyroscope measurements from the IMU to compute the ego-velocity of the vehicle. The velocity estimates are integrated through time to compute relative displacement from scan to scan. This method produces full $SE(3)$ state estimates, with the $SE(2)$ components evaluated for the $SE(2)$ benchmarks.
\end{itemize}

This collection of algorithms tests much of the released data (radar, Velodyne, Aeva, IMU), while also highlighting new challenges that the data poses.
Numerical results for $SE(2)$ odometry are presented in Table~\ref{tab:se2_odo_summary}, while results for $SE(3)$ odometry are presented in Table~\ref{tab:se3_odo_summary}.
Results are run on all 60 released sequences, but are shown as averages per route for clarity.
Odometry performance is quantified in a KITTI-style drift metric \citep{geiger2012kitti} from subsequences of length (100, 200, \dots, 800) $\si{\meter}$.
Translational results are reported as a percent drift of the subsequence length, while rotational results are reported as the amount of rotational drift accumulated over $100 \; \si{\meter}$.
This is the same odometry performance evaluation that is provided in the devkit and is used for the public leaderboard.
An additional metric is the `success rate' of each algorithm on each route, where a sequence is deemed successfully completed if the resulting translational drift is below $3\%$, a value we argue represents the threshold that all modern odometry algorithms must meet to be considered functional.

\begin{table}[t]
\centering
\caption{Summary of average $SE(2)$ relative pose accuracy for each route. KITTI odometry metric reported as \textit{XX / YY}, where \textit{XX} is translation error [\%] and \textit{YY} is rotation error [$^\circ$/100\,m]. \textit{Success} indicates the number of sequences with less than a 3\% translational drift. Averages are computed only from successful sequences. Missing sequences in the D-Aeva column correspond to those collected without an Aeva sensor.}
\label{tab:se2_odo_summary}
\begin{tabularx}{\linewidth}{
    l
    *{4}{>{\centering\arraybackslash}X}
}
\toprule
 & \textbf{DRO} & \textbf{RTR} & \textbf{LTR} & \textbf{D-Aeva} \\
\midrule
\multicolumn{5}{c}{\textit{structured routes}} \\
\texttt{suburbs}       & \small 0.20/0.04 & \small 0.36/0.05 & \small 0.15/0.04 & \small 0.37/0.06 \\[-4pt]
\scriptsize\textit{success} & \scriptsize 10/10 & \scriptsize 10/10 & \scriptsize 10/10 & \scriptsize 5/5 \\
\texttt{indust.}    & \small 0.34/0.05 & \small 0.45/0.08 & \small 0.13/0.03 & \small 0.56/0.12 \\[-4pt]
\scriptsize\textit{success} & \scriptsize 5/5 & \scriptsize 5/5 & \scriptsize 5/5 & \scriptsize 5/5 \\
\texttt{urban} & \small 0.65/0.11 & \small 1.07/0.16 & \small 0.25/0.05 &  \\[-4pt]
\scriptsize\textit{success} & \scriptsize 7/7 & \scriptsize 7/7 & \scriptsize 7/7 &  \\
\rowcolor{gray!10}
average     & \small 0.40/0.07 & \small 0.63/0.10 & \small 0.18/0.04 & \small 0.47/0.09 \\
\rowcolor{gray!10}
\textit{success}  & \small 22/22 & \small 22/22 & \small 22/22 & \small 10/10 \\

\midrule
\multicolumn{5}{c}{\textit{rural routes}} \\
\texttt{forest} & \small 0.29/0.06 & \small 0.57/0.06 & \small 0.56/0.04 &  \\[-4pt]
\scriptsize\textit{success} & \scriptsize 4/4 & \scriptsize 4/4 & \scriptsize 4/4 &  \\
\texttt{farm} & \small 0.61/0.05 & \small 1.11/0.05 & \small 0.26/0.03 &  \\[-4pt]
\scriptsize\textit{success} & \scriptsize 10/10 & \scriptsize 10/10 & \scriptsize 10/10 &  \\
\rowcolor{gray!10}
average & \small 0.45/0.06 & \small 0.84/0.06 & \small 0.41/0.03 & \small  \\
\rowcolor{gray!10}
\textit{success} & \small 14/14 & \small 14/14 & \small 14/14 & \\
\midrule

\multicolumn{5}{c}{\textit{highway routes}} \\
\texttt{tunnel}        & \small 0.35/0.04 & \small 1.25/0.05 & \small 2.03/0.04 & \small 0.34/0.03 \\[-4pt]
\scriptsize\textit{success} & \scriptsize 10/10 & \scriptsize 9/10 & \scriptsize 8/10 & \scriptsize 8/10 \\
\texttt{skyway}        & \small 0.41/0.02 & \small 0.87/0.03 & \small 0.70/0.04 & \small 0.21/0.02 \\[-4pt]
\scriptsize\textit{success} & \scriptsize 5/5 & \scriptsize 1/5 & \scriptsize 5/5 & \scriptsize 5/5 \\
\texttt{regional}       & \small 0.27/0.04 & \small 0.49/0.04 & \small 0.21/0.03 & \small 0.39/0.02 \\[-4pt]
\scriptsize\textit{success} & \scriptsize 6/6 & \scriptsize 6/6 & \scriptsize 6/6 & \scriptsize 6/6 \\
\texttt{freeway}       & \small 0.46/0.04 & \small 0.95/0.06 & \small 0.24/0.04 &  \\[-4pt]
\scriptsize\textit{success} & \scriptsize 3/3 & \scriptsize 1/3 & \scriptsize 3/3 &  \\
\rowcolor{gray!10}
average    & \small 0.37/0.04 & \small 0.89/0.05 & \small 0.79/0.04 & \small 0.31/0.02 \\
\rowcolor{gray!10}
\textit{success}  & \small 24/24 & \small 17/24 & \small 22/24 & \small 19/21 \\
\midrule

\textbf{average}     & \small \textbf{0.40/0.05} & \small \textbf{0.79/0.06} & \small \textbf{0.50/0.04} & \small \textbf{0.37/0.05} \\
\textbf{\textit{success}}   & \small \textbf{60/60} & \small \textbf{53/60} & \small \textbf{58/60} & \small \textbf{29/31} \\
\bottomrule
\end{tabularx}
\end{table}

All algorithms show low odometry drift and near-perfect success rate on structured routes.
The only caveat for these routes is that all algorithms drop in average performance on the \texttt{urban} route, likely as a result of a significant number of dynamic objects in the environment.
We also note that some drop in performance is likely artificial: the urban canyon effect lowers the quality of the ground truth and thus the evaluations become less precise.

\rev{A clear drop in performance can be noted for the structure-dependent ICP-based RTR and LTR algorithms on rural and highway routes.
The rural routes contain some sections with only vegetation, presenting inconsistent landmarks, and others with only flat farmland, lacking any landmarks entirely.
Both cases make it challenging for ICP to reliably estimate displacement between frames.
The highway routes, particularly \texttt{skyway} and \texttt{tunnel}, contain clear structure throughout the entire duration.
However, the structure they contain tends to be geometrically degenerate: repeated walls with few features to break up monotony and large sections with other vehicles on the road dominating the measurements.
Both cases make it difficult for ICP to provide an accurate estimate of motion along the direction of the road.
This is true for both considered sensing modalities, although lidar overall fares better due to a higher density of measurements and its ability to pick up on smaller features in the degenerate cases.
On the other hand, the Doppler-based methods (DRO and D-Aeva) do not show as much of a drop in performance.
These algorithms make use of environment-independent velocity measurements, which are not affected by scene degeneracy.}

All algorithms, except DRO, report failures in the more challenging rural and highway routes.
The lack of such failures for DRO can likely be attributed to the fact that the algorithm was developed on preliminary Boreas-RT data, and thus its development process already accounted for different types of roads.
This is the exact intent behind the release of this dataset.

\begin{table}[t]
\centering
\caption{Summary of average $SE(3)$ relative pose accuracy for each route. KITTI odometry metric reported as \textit{XX / YY}, where \textit{XX} is translation error [\%] and \textit{YY} is rotation error [$^\circ$/100\,m]. \textit{Success} indicates the number of sequences with less than a 3\% translational drift. Averages are computed only from successful sequences. Missing sequences in the D-Aeva column correspond to those collected without an Aeva sensor.}
\label{tab:se3_odo_summary}
\begin{tabularx}{\linewidth}{
    l
    *{2}{>{\centering\arraybackslash}X}
}
\toprule
 & \textbf{LTR} & \textbf{D-Aeva} \\
\midrule

\multicolumn{3}{c}{\textit{structured routes}} \\
\texttt{suburbs}    & \small 0.29/0.09 & \small 0.68/0.18 \\[-4pt]
\scriptsize\textit{success} & \scriptsize 10/10 & \scriptsize 5/5 \\
\texttt{industr.}   & \small 0.25/0.09 & \small 0.77/0.27 \\[-4pt]
\scriptsize\textit{success} & \scriptsize 5/5 & \scriptsize 4/5 \\
\texttt{urban}      & \small 0.49/0.20 &  \\[-4pt]
\scriptsize\textit{success} & \scriptsize 7/7 &  \\
\rowcolor{gray!10}
average              & \small 0.34/0.13 & \small 0.73/0.23 \\
\rowcolor{gray!10}
\textit{success}     & \small 22/22 & \small 9/10 \\

\midrule
\multicolumn{3}{c}{\textit{rural routes}} \\
\texttt{forest}     & \small 0.70/0.11 &  \\[-4pt]
\scriptsize\textit{success} & \scriptsize 4/4 &  \\
\texttt{farm}       & \small 0.44/0.12 &  \\[-4pt]
\scriptsize\textit{success} & \scriptsize 10/10 &  \\
\rowcolor{gray!10}
average              & \small 0.57/0.11 &  \\
\rowcolor{gray!10}
\textit{success}     & \small 14/14 &  \\

\midrule
\multicolumn{3}{c}{\textit{highway routes}} \\
\texttt{tunnel}     & \small 2.06/0.10 & \small 1.14/0.23 \\[-4pt]
\scriptsize\textit{success} & \scriptsize 8/10 & \scriptsize 8/10 \\
\texttt{skyway}     & \small 0.77/0.08 & \small 1.32/0.16 \\[-4pt]
\scriptsize\textit{success} & \scriptsize 5/5 & \scriptsize 5/5 \\
\texttt{regional}    & \small 0.35/0.09 & \small 0.77/0.14 \\[-4pt]
\scriptsize\textit{success} & \scriptsize 6/6 & \scriptsize 6/6 \\
\texttt{freeway}    & \small 0.46/0.12 &  \\[-4pt]
\scriptsize\textit{success} & \scriptsize 3/3 &  \\
\rowcolor{gray!10}
average              & \small 0.91/0.10 & \small 1.08/0.18 \\
\rowcolor{gray!10}
\textit{success}     & \small 22/24 & \small 19/21 \\

\midrule
\textbf{average}     & \small \textbf{0.64/0.11} & \small \textbf{0.94/0.20} \\
\textbf{\textit{success}} & \small \textbf{58/60} & \small \textbf{28/31} \\
\bottomrule
\end{tabularx}
\end{table}

\subsection{Localization}
Benchmarked localization algorithms are presented below, and each uses gyroscope data from the DMU41 IMU.

\begin{itemize}
    \item RTR \citep{are_we_ready_for}: The localization branch of the radar teach \& repeat pipeline uses a map built during odometry and combines an odometry prior with ICP-based map matching at each radar frame. RTR performs \textit{topometric localization}, meaning that scans are localized to locally consistent submaps instead of a globally consistent map. This alleviates the need to do loop closures and other types of global alignments. Only $SE(2)$ localization estimates are possible given the 2D radar scans.
    \item LTR \citep{are_we_ready_for}: The localization branch of the LTR pipeline, with localization done as in the RTR pipeline, but in full $SE(3)$.
\end{itemize}

\begin{table}[t]
\centering
\caption{Summary of average $SE(2)$ longitudinal, lateral, and yaw RMSE localization errors from a radar teach \& repeat pipeline for each route. \textit{Success} indicates the number of sequences with longitudinal \& lateral accuracy below 3 $\si{\meter}$.}
\label{tab:se2_loc_summary}
\setlength{\tabcolsep}{3pt}
\renewcommand{\arraystretch}{1.05}
\begin{tabularx}{\columnwidth}{
    l
    *{3}{>{\centering\arraybackslash}X}
    >{\centering\arraybackslash}X
}
\toprule
 & \textbf{long. [\si{\meter}]} & \textbf{lat. [\si{\meter}]} & \textbf{yaw [$^\circ$]} & \textbf{success} \\
\midrule

\multicolumn{5}{c}{\textit{structured routes}} \\
\texttt{suburbs}    & \small 0.101 & \small 0.066 & \small 0.116 & 9/9 \\
\texttt{industr.}   & \small 0.067 & \small 0.057 & \small 0.110 & 4/4 \\
\rowcolor{gray!10}[\tabcolsep]
average             & \small 0.084 & \small 0.062 & \small 0.113 & 13/13 \\

\midrule
\multicolumn{5}{c}{\textit{rural routes}} \\
\texttt{forest}     & \small 0.430 & \small 0.361 & \small 0.099 & 3/3 \\
\texttt{farm}       & \small 0.254 & \small 0.398 & \small 0.231 & 5/9 \\
\rowcolor{gray!10}[\tabcolsep]
average             & \small 0.342 & \small 0.379 & \small 0.165 & 8/12 \\

\midrule
\multicolumn{5}{c}{\textit{highway routes}} \\
\texttt{tunnel}     & \small 0.887 & \small 0.097 & \small 0.119 & 6/8 \\
\texttt{skyway}     & \small \xmark & \small \xmark & \small \xmark & 0/4 \\
\texttt{regional}    & \small 0.265 & \small 0.071 & \small 0.107 & 4/4 \\
\texttt{freeway}    & \small \xmark & \small \xmark & \small \xmark & 0/1 \\
\rowcolor{gray!10}[\tabcolsep]
average             & \small 0.576 & \small 0.084 & \small 0.113 & 10/17 \\

\midrule
\textbf{average}    & \small \textbf{0.334} & \small \textbf{0.175} & \small \textbf{0.130} & \small \textbf{31/42} \\
\bottomrule
\end{tabularx}
\end{table}

Numerical results for $SE(2)$ localization are shown in Table~\ref{tab:se2_loc_summary}, while results for $SE(3)$ localization are shown in Table~\ref{tab:se3_loc_summary}.
For each route, the first collected sequence is used to construct the map to which all subsequent sequences are localized.
The exceptions are the `one-way' routes, for which the first sequence in each direction is used to construct a map and all sequences going the same direction are localized relative to their respective map.
The reported average is the average from both directions.
The \texttt{freeway} route, for which only one direction had more than one traversal, only uses that direction for localization.
We do not test mapping and localizing using sequences driven in the opposite direction to one another, although we encourage the study of the impact of such a setup.

Localization performance is quantified by the root mean square error (RMSE) between the estimated scan-to-submap transform and the corresponding ground-truth transform, computed separately for each component: longitudinal, lateral, vertical, roll, pitch, and yaw.
This is the same localization performance evaluation that is provided in the devkit and is used for the public leaderboard.
Localization success is determined by whether an algorithm achieves longitudinal and lateral errors below $3 \; \si{\meter}.$

The localization results follow a similar trend to the odometry results: testing on the less structured rural and highway routes, which were not directly considered during algorithm development, yields a large number of localization failures and decreased performance in successes.
Radar-based $SE(2)$ localization has a failure rate of over $26 \%$, while the lidar-based $SE(3)$ localization has a failure rate of over $33 \%$ across the considered sequences, despite the fact that both produce excellent results on the \texttt{suburbs} and \texttt{industrial} routes.
Localization inherently relies on scene geometry and is thus more affected by geometrically challenging routes.
This highlights the need for further research in AV localization, with future work considering a broader range of road types and varying road conditions.

\begin{table*}[t]
\centering
\caption{Summary of average $SE(3)$ longitudinal, lateral, vertical, roll, pitch, and yaw RMSE localization errors from a lidar teach \& repeat pipeline for each route. \textit{Success} indicates the number of sequences with longitudinal \& lateral accuracy below 3 $\si{\meter}$.}
\label{tab:se3_loc_summary}
\begin{tabularx}{\linewidth}{
    l
    *{6}{>{\centering\arraybackslash}X}
    >{\centering\arraybackslash}X
}
\toprule
 & \textbf{long. [\si{\meter}]} & \textbf{lat. [\si{\meter}]} & \textbf{vert. [\si{\meter}]} &
   \textbf{roll [$^\circ$]} & \textbf{pitch [$^\circ$]} & \textbf{yaw [$^\circ$]} &
   \textbf{success} \\
\midrule

\multicolumn{8}{c}{\textit{structured routes}} \\
\texttt{suburbs}    & \small 0.037 & \small 0.020 & \small 0.119 & \small 0.092 & \small 0.044 & \small 0.031 & 9/9 \\
\texttt{indust.}   & \small 0.025 & \small 0.017 & \small 0.031 & \small 0.016 & \small 0.013 & \small 0.024 & 4/4 \\
\rowcolor{gray!10}[\tabcolsep]
average             & \small 0.031 & \small 0.019 & \small 0.075 & \small 0.054 & \small 0.029 & \small 0.028 & 13/13 \\

\midrule
\multicolumn{8}{c}{\textit{rural routes}} \\
\texttt{forest}     & \small 0.374 & \small 0.245 & \small 0.278 & \small 0.037 & \small 0.018 & \small 0.027 & 1/3 \\
\texttt{farm}       & \small 0.405 & \small 0.037 & \small 0.071 & \small 0.065 & \small 0.032 & \small 0.043 & 9/9 \\
\rowcolor{gray!10}[\tabcolsep]
average             & \small 0.390 & \small 0.141 & \small 0.175 & \small 0.051 & \small 0.025 & \small 0.035 & 10/12 \\

\midrule
\multicolumn{8}{c}{\textit{highway routes}} \\
\texttt{tunnel}     & \small \xmark & \small \xmark & \small \xmark & \small \xmark & \small \xmark & \small \xmark & 0/8 \\
\texttt{skyway}     & \small \xmark & \small \xmark & \small \xmark & \small \xmark & \small \xmark & \small \xmark & 0/4 \\
\texttt{regional}    & \small 0.031 & \small 0.018 & \small 0.027 & \small 0.017 & \small 0.012 & \small 0.024 & 4/4 \\
\texttt{freeway}    & \small 0.052 & \small 0.030 & \small 0.055 & \small 0.053 & \small 0.023 & \small 0.042 & 1/1 \\
\rowcolor{gray!10}[\tabcolsep]
average             & \small 0.042 & \small 0.024 & \small 0.041 & \small 0.035 & \small 0.018 & \small 0.033 & 5/17 \\

\midrule
\textbf{average}    & \small \textbf{0.154} & \small \textbf{0.061} & \small \textbf{0.097} & \small \textbf{0.047} & \small \textbf{0.024} & \small \textbf{0.032} & \small \textbf{28/42} \\
\bottomrule
\end{tabularx}
\end{table*}

\begin{figure}[t]
    \centering
    \includegraphics[width=0.48\textwidth, trim=0 115px 0 115px, clip]{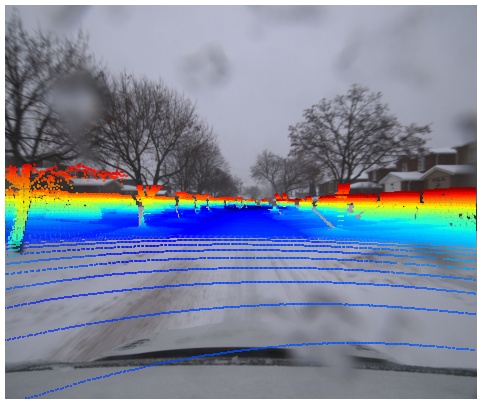}
    \hfill
    \includegraphics[width=0.48\textwidth, trim=0 115px 0 115px, clip]{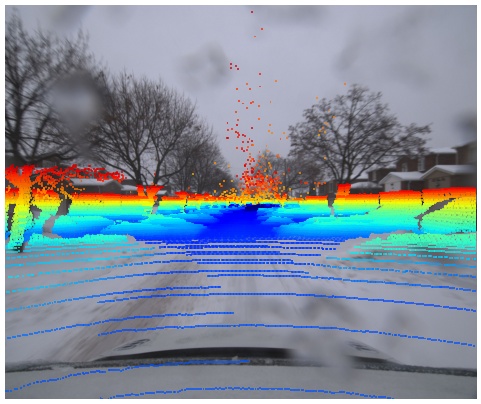}
    \caption{Lidar pointclouds from the same location on a clear day (top) and a snowy day (bottom) projected onto a camera image of the location during the snowy day. Pointclouds are coloured by elevation. Snowbanks cover the ground in the snowy day resulting in elevated $z$ axis localization errors.}
    \label{fig:snow_banks}
\end{figure}

\subsubsection{An aside on snowbanks:}
The $SE(3)$ localization benchmark on the \texttt{suburbs} route performs notably worse than reported in the original \cite{are_we_ready_for} publication that introduced this approach. Further investigation revealed that while half of the evaluated sequences matched the performance reported in the original paper, the remaining sequences exhibited notably worse performance, particularly in the vertical, roll, and pitch components.
This degradation was associated with data collected after a significant snowstorm, which resulted in substantial snowbanks along the roadside.
Figure~\ref{fig:snow_banks} visualizes the accumulated snowbanks from a lidar scan.
Because the data used to construct the localization maps was collected on a clear day, the snowbanks introduced an inconsistent, but significant, vertical offset relative to the expected road ground plane.
This offset led to degraded alignment of the lidar pointclouds and, consequently, reduced localization accuracy, particularly in the vertical, roll, and pitch directions.

Table~\ref{tab:se3_snowbanks_loc} reports \texttt{suburbs} localization results stratified by whether the data used for map construction and localization was collected on a clear day or on a day with snowbanks.
When both the map and localization data are collected on clear days or both collected on days with snowbanks, performance reaches SOTA levels.
In contrast, mixing clear-day and snowbank data leads to performance degradation in all cases.
This investigation underscores the continued need to evaluate localization algorithms across a wide range of road conditions (e.g., weather, traffic, and time of day) as well as diverse road types.

\begin{table*}[t]
\centering
\caption{Summary of average $SE(3)$ localization errors from a lidar teach \& repeat pipeline on the \texttt{suburbs} route, isolating the impact of snowbanks. Results are grouped by map (clear vs.\ snowbanks) and localization conditions (clear vs.\ snowbanks). Performance degrades under mismatched conditions but remains at SOTA levels otherwise. \textit{Success} indicates the number of sequences with longitudinal \& lateral accuracy below 3 $\si{\meter}$.}
\label{tab:se3_snowbanks_loc}
\begin{tabularx}{\linewidth}{
    l
    *{6}{>{\centering\arraybackslash}X}
    >{\centering\arraybackslash}X
}
\toprule
 & \textbf{long. [\si{\meter}]} & \textbf{lat. [\si{\meter}]} & \textbf{vert. [\si{\meter}]} &
   \textbf{roll [$^\circ$]} & \textbf{pitch [$^\circ$]} & \textbf{yaw [$^\circ$]} &
   \textbf{success} \\

\midrule
\multicolumn{8}{c}{\textit{map: clear conditions}} \\
clear conditions & \small 0.033 & \small 0.014 & \small 0.030 & \small 0.016 & \small 0.011 & \small 0.023 & 4/4 \\
snowbanks  & \small 0.040 & \small 0.025 & \small 0.189 & \small 0.153 & \small 0.072 & \small 0.038 & 5/5 \\
\rowcolor{gray!10}[\tabcolsep]
average & \small 0.037 & \small 0.020 & \small 0.119 & \small 0.092 & \small 0.044 & \small 0.031 & 9/9 \\

\midrule
\multicolumn{8}{c}{\textit{map: snowbanks}} \\
clear conditions & \small 0.044 & \small 0.022 & \small 0.171 & \small 0.205 & \small 0.085 & \small 0.050 & 5/5 \\
snowbanks & \small 0.043 & \small 0.021 & \small 0.056 & \small 0.051 & \small 0.027 & \small 0.033 & 4/4 \\
\rowcolor{gray!10}[\tabcolsep]
average & \small 0.044 & \small 0.022 & \small 0.120 & \small 0.137 & \small 0.059 & \small 0.043 & 9/9 \\

\bottomrule
\end{tabularx}
\end{table*}
\section{Development Kit}
The Boreas development kit \citep{burnett2023boreas} is extended to work with additional streams of data from IMUs, the wheel encoder, and the Aeva lidar.
The devkit provides a Python wrapper for the dataset to make it easier for new users to load in raw data and ground truth, and to evaluate odometry and localization performance.
We also provide some common methods such as pointcloud motion undistortion, radar polar image to Cartesian image unwrapping, and different visualization tools.
Several tutorials are provided as Jupyter notebooks.
The development kit is available at \href{https://www.boreas.utias.utoronto.ca/#/boreasRT}{boreas.utias.utoronto.ca}
\section{Conclusion}
This paper presents a new dataset covering repeated traversals of nine different routes, chosen for the diversity and challenges that they represent to SOTA AV state-estimation tasks of odometry, mapping, and localization.
The dataset is split up into 60 sequences totalling 643 $\si{\km}$ of on-road data.
The data contains measurements made by a 5MP FLIR camera, a $360 \si{\degree}$ Navtech radar, a $360 \si{\degree}$ Velodyne lidar, an FMCW Aeva lidar, a stand-alone Silicon Sensing IMU, and a wheel encoder.
This dataset aims to provide both a breadth and depth of testing conditions for state estimation algorithms: repeated traversals using a large collection of sensors, a variety of road types, and different driving conditions for each road type.
We provide benchmarks showcasing the types of failures that current SOTA algorithms experience on this new data.
To facilitate the future resolution of these failures, we provide a devkit and a public leaderboard on which researchers can compare their algorithms.

\begin{acks}
Thank you to Keenan Burnett, and all other authors on the original Boreas dataset, for their hard work in setting up the Boreas data collection platform, original development kit, and original website implementation.
Their well thought-through design made it much easier to extend the dataset.
Thank you also to the Navtech team for their help in setting up the firmware to enable the Doppler mode on the radar.
Thank you to Silicon Sensing for the subsidized DMU41 sensor.
This work was supported by the PGS D scholarship provided by the Natural Sciences and Engineering Research Council (NSERC) of Canada.
The Amazon Open Data Sponsorship program supports this project by hosting the Boreas and Boreas-RT datasets.

\end{acks}

\bibliographystyle{SageH}
\bibliography{references}

\clearpage
\onecolumn
\appendix
\section{Appendix}
\subsection{Full Benchmark Results}

\begin{longtable}{@{}L L *{4}{C}@{}}
\caption{Summary of $SE(2)$ relative pose accuracy on all sequences. KITTI odometry metric reported as \textit{XX / YY}, where \textit{XX} is translation error [\%] and \textit{YY} is rotation error [$^\circ$/100\,m]. Failures, results with above a 3\% translational drift, are indicated by a \xmark. Averages are computed only from successful sequences. Missing sequences in the D-Aeva column correspond to those collected without an Aeva sensor.}
\label{tab:se2_odo}\\
\toprule
\textbf{sequence type} & \textbf{sequence ID} & \textbf{DRO} & \textbf{RTR} w/ gyro & \textbf{LTR} w/ gyro & \textbf{D-Aeva} \\
\midrule
\endfirsthead

\toprule
\textbf{sequence type} & \textbf{sequence ID} & \textbf{DRO} & \textbf{RTR} w/ gyro & \textbf{LTR} w/ gyro & \textbf{D-Aeva} \\
\midrule
\endhead

\texttt{suburbs} & \scriptsize{2024-12-03-12-54} & 0.20/0.05 & 0.39/0.07 & 0.15/0.04 & 0.63/0.13 \\
& \scriptsize{2024-12-05-14-25} & 0.20/0.05 & 0.38/0.06 & 0.17/0.04 & 0.26/0.03 \\ 
& \scriptsize{2025-01-08-10-59} & 0.18/0.04 & 0.39/0.06 & 0.14/0.04 & 0.32/0.07 \\ 
& \scriptsize{2025-01-08-11-22} & 0.14/0.03 & 0.27/0.04 & 0.14/0.03 & 0.35/0.03 \\ 
& \scriptsize{2025-01-08-12-28} & 0.25/0.02 & 0.32/0.04 & 0.12/0.03 & 0.31/0.03 \\ 
& \scriptsize{2025-02-15-16-58} & 0.18/0.03 & 0.33/0.04 & 0.13/0.03 &  \\ 
& \scriptsize{2025-02-15-17-19} & 0.21/0.03 & 0.38/0.04 & 0.12/0.03 &  \\ 
& \scriptsize{2025-02-21-14-51} & 0.26/0.04 & 0.41/0.06 & 0.17/0.04 &  \\ 
& \scriptsize{2025-02-22-11-32} & 0.15/0.03 & 0.39/0.05 & 0.15/0.04 &  \\ 
& \scriptsize{2025-02-22-12-26} & 0.21/0.04 & 0.30/0.04 & 0.16/0.04 &  \\
\cmidrule(lr){3-6}
& & \textbf{0.20/0.04} & \textbf{0.36/0.05} & \textbf{0.15/0.04} & \textbf{0.37/0.06} \\
\midrule

\texttt{regional} & \scriptsize{2024-12-03-13-13} & 0.35/0.06 & 0.73/0.06 & 0.22/0.04 & 0.42/0.02 \\
& \scriptsize{2024-12-03-13-34} & 0.21/0.04 & 0.44/0.04 & 0.21/0.04 & 0.41/0.02 \\
& \scriptsize{2024-12-10-12-07} & 0.27/0.04 & 0.47/0.05 & 0.19/0.03 & 0.43/0.02 \\
& \scriptsize{2024-12-10-12-24} & 0.16/0.03 & 0.26/0.03 & 0.20/0.03 & 0.35/0.01 \\
& \scriptsize{2024-12-10-12-38} & 0.25/0.04 & 0.60/0.04 & 0.19/0.03 & 0.40/0.05 \\
& \scriptsize{2024-12-10-12-56} & 0.40/0.03 & 0.48/0.03 & 0.22/0.03 & 0.34/0.02 \\
\cmidrule(lr){3-6}
& & \textbf{0.27/0.04} & \textbf{0.49/0.04} & \textbf{0.21/0.03} & \textbf{0.39/0.02} \\
\midrule

\texttt{tunnel}& \scriptsize{2024-12-04-14-28} & 0.28/0.06 & 1.04/0.07 & 2.22/0.04 & 0.32/0.03 \\
& \scriptsize{2024-12-04-14-34} & 0.50/0.05 & \xmark & \xmark & 0.39/0.03 \\
& \scriptsize{2024-12-04-14-38} & 0.22/0.03 & 1.29/0.03 & 2.78/0.04 & 0.25/0.01 \\
& \scriptsize{2024-12-04-14-44} & 0.59/0.04 & 1.09/0.05 & 1.68/0.04 & 0.32/0.02 \\
& \scriptsize{2024-12-04-14-50} & 0.15/0.02 & 1.31/0.03 & \xmark & 0.28/0.02 \\
& \scriptsize{2024-12-04-14-59} & 0.35/0.05 & 1.21/0.07 & 2.04/0.04 & 0.55/0.02 \\
& \scriptsize{2024-12-04-15-04} & 0.18/0.03 & 1.11/0.03 & 2.30/0.03 & 0.33/0.03 \\
& \scriptsize{2024-12-04-15-10} & 0.51/0.04 & 2.26/0.07 & 0.98/0.04 & \xmark \\
& \scriptsize{2024-12-04-15-19} & 0.26/0.05 & 0.90/0.04 & 2.82/0.04 & 0.26/0.04 \\
& \scriptsize{2024-12-04-15-24} & 0.48/0.03 & 1.07/0.04 & 1.39/0.05 & \xmark \\
\cmidrule(lr){3-6}
& & \textbf{0.35/0.04} & \textbf{1.25/0.05} & \textbf{2.03/0.04} & \textbf{0.34/0.03} \\
\midrule

\texttt{skyway} & \scriptsize{2024-12-04-11-45} & 0.34/0.02 & \xmark & 1.56/0.04 & 0.20/0.01 \\
& \scriptsize{2024-12-04-11-56} & 0.61/0.03 & \xmark & 0.40/0.04 & 0.23/0.02 \\
& \scriptsize{2024-12-04-12-08} & 0.28/0.02 & 0.87/0.03 & 0.57/0.04 & 0.23/0.02 \\
& \scriptsize{2024-12-04-12-19} & 0.44/0.02 & \xmark & 0.44/0.05 & 0.22/0.01 \\
& \scriptsize{2024-12-04-12-34} & 0.37/0.02 & \xmark & 0.52/0.04 & 0.19/0.02 \\
\cmidrule(lr){3-6}
& & \textbf{0.41/0.02} & \textbf{0.87/0.03} & \textbf{0.70/0.04} & \textbf{0.21/0.02} \\
\midrule

\texttt{industrial} & \scriptsize{2024-12-05-14-12} & 0.16/0.03 & 0.31/0.05 & 0.11/0.03 & 1.01/0.31 \\
& \scriptsize{2024-12-23-16-27} & 0.45/0.05 & 0.56/0.12 & 0.15/0.03 & 0.42/0.04 \\
& \scriptsize{2024-12-23-16-44} & 0.49/0.05 & 0.56/0.08 & 0.14/0.03 & 0.33/0.04 \\
& \scriptsize{2024-12-23-17-01} & 0.23/0.04 & 0.34/0.08 & 0.14/0.03 & 0.60/0.14 \\
& \scriptsize{2024-12-23-17-18} & 0.35/0.05 & 0.48/0.07 & 0.13/0.03 & 0.43/0.09 \\
\cmidrule(lr){3-6}
& & \textbf{0.34/0.05} & \textbf{0.45/0.08} & \textbf{0.13/0.03} & \textbf{0.56/0.12} \\
\midrule

\texttt{forest} & \scriptsize{2025-07-18-10-33} & 0.27/0.05 & 0.56/0.06 & 0.51/0.07 &  \\
& \scriptsize{2025-07-18-11-00} & 0.27/0.05 & 0.55/0.06 & 0.58/0.02 &  \\
& \scriptsize{2025-07-18-11-25} & 0.31/0.06 & 0.59/0.06 & 0.51/0.03 &  \\
& \scriptsize{2025-07-18-11-53} & 0.31/0.06 & 0.59/0.07 & 0.64/0.03 &  \\
\cmidrule(lr){3-6}
& & \textbf{0.29/0.06} & \textbf{0.57/0.06} & \textbf{0.56/0.04} &  \\
\midrule

\texttt{farm} & \scriptsize{2025-07-18-14-55} & 0.67/0.05 & 0.76/0.05 & 0.25/0.03 &  \\
& \scriptsize{2025-07-18-15-12} & 1.20/0.04 & 1.51/0.05 & 0.27/0.02 &  \\
& \scriptsize{2025-07-18-15-30} & 1.01/0.05 & 1.34/0.05 & 0.25/0.02 &  \\
& \scriptsize{2025-07-18-15-48} & 0.74/0.05 & 2.43/0.06 & 0.44/0.03 &  \\
& \scriptsize{2025-07-18-16-05} & 0.40/0.04 & 1.21/0.05 & 0.32/0.02 &  \\
& \scriptsize{2025-08-13-09-01} & 0.40/0.06 & 0.77/0.05 & 0.22/0.03 &  \\
& \scriptsize{2025-08-13-09-21} & 0.45/0.07 & 0.77/0.07 & 0.21/0.03 &  \\
& \scriptsize{2025-08-13-09-46} & 0.44/0.06 & 0.88/0.06 & 0.21/0.03 &  \\
& \scriptsize{2025-08-13-10-12} & 0.44/0.05 & 0.75/0.05 & 0.21/0.03 &  \\
& \scriptsize{2025-08-13-10-36} & 0.39/0.05 & 0.72/0.05 & 0.19/0.03 &  \\
\cmidrule(lr){3-6}
& & \textbf{0.61/0.05} & \textbf{1.11/0.05} & \textbf{0.26/0.03} &  \\
\midrule

\texttt{freeway} & \scriptsize{2025-07-18-16-24} & 0.46/0.05 & 0.95/0.06 & 0.31/0.04 &  \\
& \scriptsize{2025-08-13-07-54} & 0.35/0.04 & \xmark & 0.21/0.04 &  \\
& \scriptsize{2025-08-13-11-52} & 0.57/0.04 & \xmark & 0.20/0.04 &  \\
\cmidrule(lr){3-6}
& & \textbf{0.46/0.04} & \textbf{0.95/0.06} & \textbf{0.24/0.04} &  \\
\midrule

\texttt{urban} & \scriptsize{2025-08-06-06-33} & 0.35/0.07 & 0.62/0.10 & 0.20/0.05 &  \\
& \scriptsize{2025-08-06-07-05} & 0.48/0.08 & 1.55/0.11 & 0.23/0.04 &  \\
& \scriptsize{2025-08-06-07-41} & 0.52/0.11 & 0.92/0.14 & 0.22/0.05 &  \\
& \scriptsize{2025-08-06-08-35} & 0.70/0.13 & 1.33/0.21 & 0.36/0.06 &  \\
& \scriptsize{2025-08-06-10-48} & 0.47/0.11 & 0.68/0.13 & 0.19/0.05 &  \\
& \scriptsize{2025-08-06-11-32} & 0.95/0.13 & 1.15/0.19 & 0.31/0.06 &  \\
& \scriptsize{2025-08-06-12-20} & 1.11/0.14 & 1.24/0.22 & 0.21/0.06 &  \\
\cmidrule(lr){3-6}
& & \textbf{0.65/0.11} & \textbf{1.07/0.16} & \textbf{0.25/0.05} &  \\
\bottomrule
\end{longtable}
\twocolumn

\onecolumn
\begin{longtable}{@{}L L *{2}{C}@{}}
\caption{Summary of $SE(3)$ relative pose accuracy on all sequences. KITTI odometry metric reported as \textit{XX / YY}, where \textit{XX} is translation error [\%] and \textit{YY} is rotation error [$^\circ$/100\,m]. Failures, results with above a 3\% translational drift, are indicated by a \xmark. Averages are computed only from successful sequences. Missing sequences in the D-Aeva column correspond to those collected without an Aeva sensor.}
\label{tab:se3_odo}\\
\toprule
\textbf{sequence type} & \textbf{ID} & \textbf{LTR w/ gyro} & \textbf{D-Aeva} \\
\midrule
\endfirsthead

\toprule
\textbf{sequence type} & \textbf{ID} & \textbf{LTR w/ gyro} & \textbf{D-Aeva} \\
\midrule
\endhead

\texttt{suburbs}
& \scriptsize{2024-12-03-12-54} & 0.30/0.10 & 1.00/0.24 \\
& \scriptsize{2024-12-05-14-25} & 0.29/0.09 & 0.52/0.18 \\
& \scriptsize{2025-01-08-10-59} & 0.30/0.09 & 0.62/0.20 \\
& \scriptsize{2025-01-08-11-22} & 0.28/0.09 & 0.64/0.14 \\
& \scriptsize{2025-01-08-12-28} & 0.26/0.08 & 0.60/0.16 \\
& \scriptsize{2025-02-15-16-58} & 0.27/0.09 &  \\
& \scriptsize{2025-02-15-17-19} & 0.27/0.09 &  \\
& \scriptsize{2025-02-21-14-51} & 0.34/0.10 &  \\
& \scriptsize{2025-02-22-11-32} & 0.29/0.09 &  \\
& \scriptsize{2025-02-22-12-26} & 0.30/0.10 &  \\
\cmidrule(lr){3-4}
& & \textbf{0.29/0.09} & \textbf{0.68/0.18} \\
\midrule

\texttt{regional} 
& \scriptsize{2024-12-03-13-13} & 0.34/0.09 & 0.77/0.15 \\
& \scriptsize{2024-12-03-13-34} & 0.34/0.09 & 0.75/0.13 \\
& \scriptsize{2024-12-10-12-07} & 0.34/0.09 & 0.79/0.13 \\
& \scriptsize{2024-12-10-12-24} & 0.35/0.10 & 0.73/0.14 \\
& \scriptsize{2024-12-10-12-38} & 0.32/0.08 & 0.82/0.16 \\
& \scriptsize{2024-12-10-12-56} & 0.38/0.10 & 0.73/0.14 \\
\cmidrule(lr){3-4}
& & \textbf{0.35/0.09} & \textbf{0.77/0.14} \\
\midrule

\texttt{tunnel}
& \scriptsize{2024-12-04-14-28} & 2.26/0.12 & 1.18/0.26 \\
& \scriptsize{2024-12-04-14-34} & \xmark & 0.80/0.33 \\
& \scriptsize{2024-12-04-14-38} & 2.82/0.11 & 1.05/0.17 \\
& \scriptsize{2024-12-04-14-44} & 1.69/0.08 & 1.61/0.25 \\
& \scriptsize{2024-12-04-14-50} & \xmark & 0.87/0.15 \\
& \scriptsize{2024-12-04-14-59} & 2.05/0.09 & 1.31/0.21 \\
& \scriptsize{2024-12-04-15-04} & 2.36/0.11 & 1.00/0.24 \\
& \scriptsize{2024-12-04-15-10} & 1.02/0.09 & \xmark \\
& \scriptsize{2024-12-04-15-19} & 2.87/0.11 & 1.29/0.22 \\
& \scriptsize{2024-12-04-15-24} & 1.41/0.09 & \xmark \\
\cmidrule(lr){3-4}
& & \textbf{2.06/0.10} & \textbf{1.14/0.23} \\
\midrule

\texttt{skyway} 
& \scriptsize{2024-12-04-11-45} & 1.62/0.08 & 1.28/0.16 \\
& \scriptsize{2024-12-04-11-56} & 0.47/0.07 & 1.28/0.19 \\
& \scriptsize{2024-12-04-12-08} & 0.63/0.08 & 1.37/0.15 \\
& \scriptsize{2024-12-04-12-19} & 0.51/0.08 & 1.39/0.16 \\
& \scriptsize{2024-12-04-12-34} & 0.60/0.08 & 1.27/0.16 \\
\cmidrule(lr){3-4}
& & \textbf{0.77/0.08} & \textbf{1.32/0.16} \\
\midrule

\texttt{industrial} 
& \scriptsize{2024-12-05-14-12} & 0.23/0.08 & \xmark \\
& \scriptsize{2024-12-23-16-27} & 0.27/0.10 & 0.84/0.28 \\
& \scriptsize{2024-12-23-16-44} & 0.24/0.09 & 0.64/0.24 \\
& \scriptsize{2024-12-23-17-01} & 0.26/0.09 & 0.86/0.31 \\
& \scriptsize{2024-12-23-17-18} & 0.24/0.09 & 0.74/0.25 \\
\cmidrule(lr){3-4}
& & \textbf{0.25/0.09} & \textbf{0.77/0.27} \\
\midrule


\texttt{forest} 
& \scriptsize{2025-07-18-10-33} & 0.66/0.14 &  \\
& \scriptsize{2025-07-18-11-00} & 0.71/0.10 &  \\
& \scriptsize{2025-07-18-11-25} & 0.64/0.10 &  \\
& \scriptsize{2025-07-18-11-53} & 0.78/0.10 &  \\
\cmidrule(lr){3-4}
& & \textbf{0.70/0.11} &  \\
\midrule

\texttt{farm} 
& \scriptsize{2025-07-18-14-55} & 0.45/0.11 &  \\
& \scriptsize{2025-07-18-15-12} & 0.44/0.10 &  \\
& \scriptsize{2025-07-18-15-30} & 0.42/0.11 &  \\
& \scriptsize{2025-07-18-15-48} & 0.60/0.11 &  \\
& \scriptsize{2025-07-18-16-05} & 0.49/0.10 &  \\
& \scriptsize{2025-08-13-09-01} & 0.40/0.13 &  \\
& \scriptsize{2025-08-13-09-21} & 0.40/0.13 &  \\
& \scriptsize{2025-08-13-09-46} & 0.42/0.13 &  \\
& \scriptsize{2025-08-13-10-12} & 0.42/0.13 &  \\
& \scriptsize{2025-08-13-10-36} & 0.40/0.12 &  \\
\cmidrule(lr){3-4}
& & \textbf{0.44/0.12} &  \\
\midrule

\texttt{freeway} 
& \scriptsize{2025-07-18-16-24} & 0.53/0.13 &  \\
& \scriptsize{2025-08-13-07-54} & 0.42/0.11 &  \\
& \scriptsize{2025-08-13-11-52} & 0.44/0.13 &  \\
\cmidrule(lr){3-4}
& & \textbf{0.46/0.12} &  \\
\midrule

\texttt{urban} 
& \scriptsize{2025-08-06-06-33} & 0.41/0.16 &  \\
& \scriptsize{2025-08-06-07-05} & 0.46/0.17 &  \\
& \scriptsize{2025-08-06-07-41} & 0.43/0.17 &  \\
& \scriptsize{2025-08-06-08-35} & 0.55/0.22 &  \\
& \scriptsize{2025-08-06-10-48} & 0.48/0.21 &  \\
& \scriptsize{2025-08-06-11-32} & 0.57/0.23 &  \\
& \scriptsize{2025-08-06-12-20} & 0.51/0.24 &  \\
\cmidrule(lr){3-4}
& & \textbf{0.49/0.20} &  \\
\bottomrule
\end{longtable}
\twocolumn

\onecolumn
\begin{longtable}{@{}L L L *{3}{C}@{}}
\caption{Summary of $SE(2)$ longitudinal, lateral, and yaw RMSE localization errors from a radar teach \& repeat pipeline for each sequence. Sequences that fail to achieve longitudinal \& lateral accuracy below 3 $\si{\meter}$ are marked with an \xmark.}
\label{tab:se2_loc}\\
\toprule
\textbf{sequence type} & \textbf{mapping ID} & \textbf{loc. ID} & \textbf{long.} & \textbf{lat.} & \textbf{yaw} \\
\midrule
\endfirsthead

\toprule
\textbf{sequence type} & \textbf{mapping ID} & \textbf{loc. ID} & \textbf{long.} & \textbf{lat.} & \textbf{yaw} \\
\midrule
\endhead

\texttt{suburbs} & \scriptsize{2024-12-03-12-54} & \scriptsize{2024-12-05-14-25} & 0.092 & 0.056 & 0.095 \\
& & \scriptsize{2025-01-08-10-59} & 0.096 & 0.051 & 0.112 \\
& & \scriptsize{2025-01-08-11-22} & 0.090 & 0.047 & 0.101 \\
& & \scriptsize{2025-01-08-12-28} & 0.093 & 0.061 & 0.118 \\
& & \scriptsize{2025-02-15-16-58} & 0.099 & 0.071 & 0.110 \\
& & \scriptsize{2025-02-15-17-19} & 0.119 & 0.066 & 0.121 \\
& & \scriptsize{2025-02-21-14-51} & 0.112 & 0.082 & 0.143 \\
& & \scriptsize{2025-02-22-11-32} & 0.107 & 0.084 & 0.127 \\
& & \scriptsize{2025-02-22-12-26} & 0.103 & 0.079 & 0.112 \\
\cmidrule(lr){4-6}
& & & \textbf{0.101} & \textbf{0.066} & \textbf{0.116} \\
\midrule

\texttt{regional} & \scriptsize{2024-12-03-13-13} & \scriptsize{2024-12-10-12-07} & 0.140 & 0.067 & 0.115 \\
& & \scriptsize{2024-12-10-12-38} & 0.672 & 0.067 & 0.117 \\
& \scriptsize{2024-12-03-13-34} & \scriptsize{2024-12-10-12-24} & 0.119 & 0.073 & 0.089 \\
& & \scriptsize{2024-12-10-12-56} & 0.131 & 0.076 & 0.107 \\
\cmidrule(lr){4-6}
& & & \textbf{0.265} & \textbf{0.071} & \textbf{0.107} \\
\midrule

\texttt{tunnel} & \scriptsize{2024-12-04-14-28} & \scriptsize{2024-12-04-14-38} & 2.505 & 0.182 & 0.216\\
& & \scriptsize{2024-12-04-14-50} & 1.418 & 0.113 & 0.097 \\
& & \scriptsize{2024-12-04-15-04} & 0.472 & 0.072 & 0.106 \\
& & \scriptsize{2024-12-04-15-19} & 0.175 & 0.062 & 0.093 \\
& \scriptsize{2024-12-04-14-44} & \scriptsize{2024-12-04-14-34} & \xmark & \xmark & \xmark \\
& & \scriptsize{2024-12-04-14-59} & 0.278 & 0.066 & 0.113 \\
& & \scriptsize{2024-12-04-15-10} & \xmark & \xmark & \xmark \\
& & \scriptsize{2024-12-04-15-24} & 0.474 & 0.087 & 0.086 \\
\cmidrule(lr){4-6}
& & & \textbf{0.887} & \textbf{0.097} & \textbf{0.119} \\
\midrule

\texttt{skyway} & \scriptsize{2024-12-04-11-45} & \scriptsize{2024-12-04-11-56} & \xmark & \xmark & \xmark \\
& & \scriptsize{2024-12-04-12-08} & \xmark & \xmark & \xmark \\
& & \scriptsize{2024-12-04-12-19} & \xmark & \xmark & \xmark \\
& & \scriptsize{2024-12-04-12-34} & \xmark & \xmark & \xmark \\
\cmidrule(lr){4-6}
& & & \xmark & \xmark & \xmark \\
\midrule

\texttt{industrial} & \scriptsize{2024-12-05-14-12} & \scriptsize{2024-12-23-16-27} & 0.066 & 0.058 & 0.106 \\
& & \scriptsize{2024-12-23-16-44} & 0.072 & 0.056 & 0.121 \\
& & \scriptsize{2024-12-23-17-01} & 0.063 & 0.056 & 0.109 \\
& & \scriptsize{2024-12-23-17-18} & 0.065 & 0.057 & 0.102 \\
\cmidrule(lr){4-6}
& & & \textbf{0.067} & \textbf{0.057} & \textbf{0.110} \\
\midrule

\texttt{forest} & \scriptsize{2025-07-18-10-33} & \scriptsize{2025-07-18-11-00} & 0.371 & 0.245 & 0.086 \\
& & \scriptsize{2025-07-18-11-25} & 0.329 & 0.330 & 0.101 \\
& & \scriptsize{2025-07-18-11-53} & 0.589 & 0.507 & 0.109 \\
\cmidrule(lr){4-6}
& & & \textbf{0.430} & \textbf{0.361} & \textbf{0.099} \\
\midrule

\texttt{farm} & \scriptsize{2025-07-18-14-55} & \scriptsize{2025-07-18-15-12} & \xmark & \xmark & \xmark \\
& & \scriptsize{2025-07-18-15-30} & \xmark & \xmark & \xmark \\
& & \scriptsize{2025-07-18-15-48} & \xmark & \xmark & \xmark \\
& & \scriptsize{2025-07-18-16-05} & 0.154 & 0.134 & 0.102 \\
& & \scriptsize{2025-08-13-09-01} & \xmark & \xmark & \xmark \\
& & \scriptsize{2025-08-13-09-21} & 0.234 & 0.256 & 0.263 \\
& & \scriptsize{2025-08-13-09-46} & 0.370 & 1.057 & 0.398 \\
& & \scriptsize{2025-08-13-10-12} & 0.327 & 0.284 & 0.199 \\
& & \scriptsize{2025-08-13-10-36} & 0.187 & 0.260 & 0.191 \\
\cmidrule(lr){4-6}
& & & \textbf{0.254} & \textbf{0.398} & \textbf{0.231} \\
\midrule

\texttt{freeway} & \scriptsize{2025-07-18-16-24} & \scriptsize{2025-08-13-11-52} & \xmark & \xmark & \xmark \\
\cmidrule(lr){4-6}
& & & \xmark & \xmark & \xmark \\

\bottomrule

\end{longtable}
\twocolumn

\onecolumn
\begin{longtable}{@{}L L L *{6}{M}@{}}
\caption{Summary of $SE(3)$ longitudinal, lateral, vertical, roll, pitch, and yaw RMSE localization errors from a lidar teach \& repeat pipeline for each sequence. Sequences that fail to achieve longitudinal \& lateral accuracy below 3 $\si{\meter}$ are marked with an \xmark.}
\label{tab:se3_loc}\\
\toprule
\textbf{sequence type} & \textbf{mapping ID} & \textbf{loc. ID} & \textbf{long.} & \textbf{lat.} & \textbf{vert.} & \textbf{roll} & \textbf{pitch} & \textbf{yaw} \\
\midrule
\endfirsthead

\toprule
\textbf{sequence type} & \textbf{mapping ID} & \textbf{loc. ID} & \textbf{long.} & \textbf{lat.} & \textbf{vert.} & \textbf{roll} & \textbf{pitch} & \textbf{yaw} \\
\midrule
\endhead

\texttt{suburbs} & \scriptsize{2024-12-03-12-54} & \scriptsize{2024-12-05-14-25} & 0.033 & 0.011 & 0.024 & 0.018 & 0.011 & 0.026 \\
& & \scriptsize{2025-01-08-10-59} & 0.043 & 0.017 & 0.045 & 0.016 & 0.011 & 0.025 \\
& & \scriptsize{2025-01-08-11-22} & 0.026 & 0.012 & 0.021 & 0.015 & 0.011 & 0.019 \\
& & \scriptsize{2025-01-08-12-28} & 0.029 & 0.016 & 0.030 & 0.016 & 0.011 & 0.020 \\
& & \scriptsize{2025-02-15-16-58} & 0.032 & 0.020 & 0.190 & 0.121 & 0.053 & 0.029 \\
& & \scriptsize{2025-02-15-17-19} & 0.032 & 0.020 & 0.193 & 0.117 & 0.059 & 0.030 \\
& & \scriptsize{2025-02-21-14-51} & 0.058 & 0.026 & 0.187 & 0.175 & 0.078 & 0.045 \\
& & \scriptsize{2025-02-22-11-32} & 0.046 & 0.034 & 0.191 & 0.187 & 0.079 & 0.044 \\
& & \scriptsize{2025-02-22-12-26} & 0.032 & 0.027 & 0.186 & 0.163 & 0.092 & 0.042 \\
\cmidrule(lr){4-9}
& & & \textbf{0.037} & \textbf{0.020} & \textbf{0.119} & \textbf{0.092} & \textbf{0.044} & \textbf{0.031} \\
\midrule

\texttt{regional} & \scriptsize{2024-12-03-13-13} & \scriptsize{2024-12-10-12-07} & 0.033 & 0.015 & 0.019 & 0.016 & 0.012 & 0.024 \\
& & \scriptsize{2024-12-10-12-38} & 0.032 & 0.018 & 0.020 & 0.016 & 0.012 & 0.023 \\
& \scriptsize{2024-12-03-13-34} & \scriptsize{2024-12-10-12-24} & 0.028 & 0.017 & 0.042 & 0.018 & 0.011 & 0.021 \\
& & \scriptsize{2024-12-10-12-56} & 0.031 & 0.022 & 0.027 & 0.017 & 0.011 & 0.027 \\
\cmidrule(lr){4-9}
& & & \textbf{0.031} & \textbf{0.018} & \textbf{0.027} & \textbf{0.017} & \textbf{0.012} & \textbf{0.024} \\
\midrule

\texttt{tunnel}& \scriptsize{2024-12-04-14-28} & \scriptsize{2024-12-04-14-38} & \xmark & \xmark & \xmark & \xmark & \xmark & \xmark \\
& & \scriptsize{2024-12-04-14-50} & \xmark & \xmark & \xmark & \xmark & \xmark & \xmark\\
& & \scriptsize{2024-12-04-15-04} & \xmark & \xmark & \xmark & \xmark & \xmark & \xmark \\
& & \scriptsize{2024-12-04-15-19} & \xmark & \xmark & \xmark & \xmark & \xmark & \xmark\\
& \scriptsize{2024-12-04-14-44} & \scriptsize{2024-12-04-14-34} & \xmark & \xmark & \xmark & \xmark & \xmark & \xmark \\
& & \scriptsize{2024-12-04-14-59} & \xmark & \xmark & \xmark & \xmark & \xmark & \xmark \\
& & \scriptsize{2024-12-04-15-10} & \xmark & \xmark & \xmark & \xmark & \xmark & \xmark \\
& & \scriptsize{2024-12-04-15-24} & \xmark & \xmark & \xmark & \xmark & \xmark & \xmark \\
\cmidrule(lr){4-9}
& & & \xmark & \xmark & \xmark & \xmark & \xmark & \xmark \\
\midrule

\texttt{skyway} & \scriptsize{2024-12-04-11-45} & \scriptsize{2024-12-04-11-56} & \xmark & \xmark & \xmark & \xmark & \xmark & \xmark \\
& & \scriptsize{2024-12-04-12-08} & \xmark & \xmark & \xmark & \xmark & \xmark & \xmark \\
& & \scriptsize{2024-12-04-12-19} & \xmark & \xmark & \xmark & \xmark & \xmark & \xmark \\
& & \scriptsize{2024-12-04-12-34} & \xmark & \xmark & \xmark & \xmark & \xmark & \xmark \\
\cmidrule(lr){4-9}
& & & \xmark & \xmark & \xmark & \xmark & \xmark & \xmark \\
\midrule

\texttt{industrial} & \scriptsize{2024-12-05-14-12} & \scriptsize{2024-12-23-16-27} & 0.027 & 0.015 & 0.031 & 0.015 & 0.013 & 0.021 \\
& & \scriptsize{2024-12-23-16-44} & 0.022 & 0.021 & 0.030 & 0.014 & 0.012 & 0.020 \\
& & \scriptsize{2024-12-23-17-01} & 0.025 & 0.014 & 0.035 & 0.019 & 0.014 & 0.029 \\
& & \scriptsize{2024-12-23-17-18} & 0.024 & 0.017 & 0.027 & 0.017 & 0.013 & 0.028 \\
\cmidrule(lr){4-9}
& & & \textbf{0.025} & \textbf{0.017} & \textbf{0.031} & \textbf{0.016} & \textbf{0.013} & \textbf{0.024} \\
\midrule

\texttt{forest} & \scriptsize{2025-07-18-10-33} & \scriptsize{2025-07-18-11-00} & 0.374 & 0.245 & 0.278 & 0.037 & 0.018 & 0.027 \\
& & \scriptsize{2025-07-18-11-25} & \xmark & \xmark & \xmark & \xmark & \xmark & \xmark \\
& & \scriptsize{2025-07-18-11-53} & \xmark & \xmark & \xmark & \xmark & \xmark & \xmark \\
\cmidrule(lr){4-9}
& & & \textbf{0.374} & \textbf{0.245} & \textbf{0.278} & \textbf{0.037} & \textbf{0.018} & \textbf{0.027} \\
\midrule

\texttt{farm} & \scriptsize{2025-07-18-14-55} & \scriptsize{2025-07-18-15-12} 
& 0.064 & 0.027 & 0.060 & 0.016 & 0.013 & 0.025 \\
& & \scriptsize{2025-07-18-15-30} 
& 0.064 & 0.025 & 0.069 & 0.017 & 0.013 & 0.024 \\
& & \scriptsize{2025-07-18-15-48} 
& 0.148 & 0.030 & 0.076 & 0.017 & 0.070 & 0.031 \\
& & \scriptsize{2025-07-18-16-05} 
& 0.158 & 0.028 & 0.074 & 0.016 & 0.014 & 0.027 \\
& & \scriptsize{2025-08-13-09-01} 
& 0.425 & 0.041 & 0.049 & 0.106 & 0.038 & 0.069 \\
& & \scriptsize{2025-08-13-09-21} 
& 0.618 & 0.044 & 0.053 & 0.102 & 0.034 & 0.045 \\
& & \scriptsize{2025-08-13-09-46} 
& 0.662 & 0.047 & 0.078 & 0.103 & 0.035 & 0.048 \\
& & \scriptsize{2025-08-13-10-12} 
& 0.745 & 0.046 & 0.089 & 0.103 & 0.035 & 0.053 \\
& & \scriptsize{2025-08-13-10-36} 
& 0.760 & 0.045 & 0.088 & 0.101 & 0.035 & 0.062 \\
\cmidrule(lr){4-9}
& & & \textbf{0.405} & \textbf{0.037} & \textbf{0.071} & \textbf{0.065} & \textbf{0.032} & \textbf{0.043} \\
\midrule

\texttt{freeway} & \scriptsize{2025-07-18-16-24} & \scriptsize{2025-08-13-11-52} & 0.052 & 0.030 & 0.055 & 0.053 & 0.023 & 0.042 \\
\cmidrule(lr){4-9}
& & & \textbf{0.052} & \textbf{0.030} & \textbf{0.055} & \textbf{0.053} & \textbf{0.023} & \textbf{0.042} \\

\bottomrule

\end{longtable}
\twocolumn

\end{document}